\documentclass[10pt,twocolumn,letterpaper]{article}

\usepackage[pagenumbers]{cvpr} % To force page numbers, e.g. for an arXiv version
\usepackage{multicol}
\usepackage{multirow}
\usepackage{diagbox}
\usepackage{colortbl}
\usepackage{float}
\definecolor{cvprblue}{rgb}{0.21,0.49,0.74}
\usepackage[pagebackref,breaklinks,colorlinks,allcolors = cvprblue]{hyperref}

\def\paperID{824} % *** Enter the Paper ID here
\def\confName{CVPR}
\def\confYear{2025}

\title{Anomize: Better Open Vocabulary Video Anomaly Detection}
\author{
Fei Li$^{1,2\#}$ \quad
Wenxuan Liu$^{3,4\#}$ \quad
Jingjing Chen$^{2}$ \quad
Ruixu Zhang$^{1}$ \quad
Yuran Wang$^{1}$ \\
Xian Zhong$^{4}$ \quad
Zheng Wang$^{1}$\thanks{\ Corresponding author. \# Contributed equally to this work.} \\
\small
$^1$National Engineering Research Center for Multimedia Software, School of Computer Science, Wuhan University 
\\
\small
$^2$Shanghai Key Lab of Intelligent Information Processing, School of Computer Science, Fudan University \\
\small
$^3$State Key Laboratory for Multimedia Information Processing, School of Computer Science, Peking University \\
\small
$^4$Hubei Key Laboratory of Transportation Internet of Things, 
Wuhan University of Technology \\
{\tt\small lifeiwhu@whu.edu.cn, liuwx66@pku.edu.cn}
}
\begin{document}
% \captionsetup{skip = 7pt}
% \setlength{\textfloatsep}{5pt}
\maketitle

\begin{abstract}

Open Vocabulary Video Anomaly Detection (OVVAD) seeks to detect and classify both base and novel anomalies. However, existing methods face two specific challenges related to novel anomalies. The first challenge is detection ambiguity, where the model struggles to assign accurate anomaly scores to unfamiliar anomalies. The second challenge is categorization confusion, where novel anomalies are often misclassified as visually similar base instances. To address these challenges, we explore supplementary information from multiple sources to mitigate detection ambiguity by leveraging multiple levels of visual data alongside matching textual information. Furthermore, we propose incorporating label relations to guide the encoding of new labels, thereby improving alignment between novel videos and their corresponding labels, which helps reduce categorization confusion. The resulting Anomize framework effectively tackles these issues, achieving superior performance on \textsc{UCF-Crime} and \textsc{XD-Violence} datasets, demonstrating its effectiveness in OVVAD.

\end{abstract}

\section{Introduction}
\label{sec:intro}
% 介绍异常检测任务的重要性 + 开放世界的必要性(以前的任务主要是半监督和弱监督这种闭集任务不适用于开放世界)。
% 具体：视频异常检测（VAD）致力于检测视频中的异常事件，在公共安全和智能监控中有着广泛的应用。传统的VAD根据监督模式可以大致分为两种类型，即半监督VAD和弱监督VAD。半监督VAD假设在训练阶段只有正常样本可用，不符合这些正常训练样本的测试样本被识别为异常。弱监督VAD可以被视为一个二元分类任务，假设在训练阶段正常和异常样本都可用，但异常事件的精确时间注释是未知的。这些异常检测都是闭集任务，即仅限于检测一组封闭的异常类别，无法处理任意的不可见的异常。这种限制限制了它们在开放世界场景中的应用，并造成了增加报告丢失的风险，因为实际部署中的许多真实世界的异常情况并没有出现在训练数据中。

Video Anomaly Detection (VAD) identifies anomaly in videos and is widely used in public safety systems. Traditional VAD methods can be categorized based on the type of training data used. Semi-supervised VAD~\cite{zaheer2020old, cong2011sparse, liu2018future} is trained exclusively on normal samples, detecting anomalies as deviations from learned normal patterns. In contrast, weakly supervised VAD~\cite{sultani2018real, zaheer2020claws, joo2023clip} is trained on both normal and anomalous samples but lacks precise temporal labels, treating VAD as a binary classification problem. 
Both methods focus on detecting specific anomaly within a closed set and exhibit limitations in open-world scenarios.

Recent research has explored open-set VAD~\cite{acsintoae2022ubnormal, hirschorn2023normalizing}, where anomalies seen in training are considered base cases, while others are treated as novel cases. It trains on normal and base anomalies to detect all anomalies, overcoming the limitations of closed-set detection. However, it struggles with understanding anomaly categories, leading to unclear outputs~\cite{wu2024open}. Consequently, a leading study~\cite{wu2024open} has further investigated open vocabulary (OV) VAD, which aims to detect and categorize all anomalies using the same training data as open-set VAD, offering more informative results.

% Therefore, we focus on this task.
% However, it presents certain challenges related to novel cases, as depicted in Fig. \ref{fig:motivation1}.

% \setlength{\textfloatsep}{8pt}
% \begin{figure} 
% 	\centering
% 	\begin{tabular}{cc}
% 	\includegraphics[width = 0.43 \linewidth]{sec/images/motivation1_2v6.pdf} & 
% 	\includegraphics[width = 0.45 \linewidth]{sec/images/motivation1_1v6.pdf} \\
% 	\small{(a) Detection Ambiguity} & \small{(b) Categorization Confusion} \\
% 	\end{tabular}
% 	\vspace{-5pt}
% 	\caption{\textbf{Challenges Related to Novel Anomalies.} 
% 	(a) Detection ambiguity: The model struggles to assign accurate anomaly scores to unfamiliar frames containing novel anomalies. 
% 	(b) Categorization confusion: Novel anomalies are misclassified as visually similar instances from the training set.}
% 	\label{fig:motivation1} 
% 	\vspace{-5pt}
% \end{figure}

\begin{figure} 
	\centering
	\includegraphics[width = 0.9\linewidth]{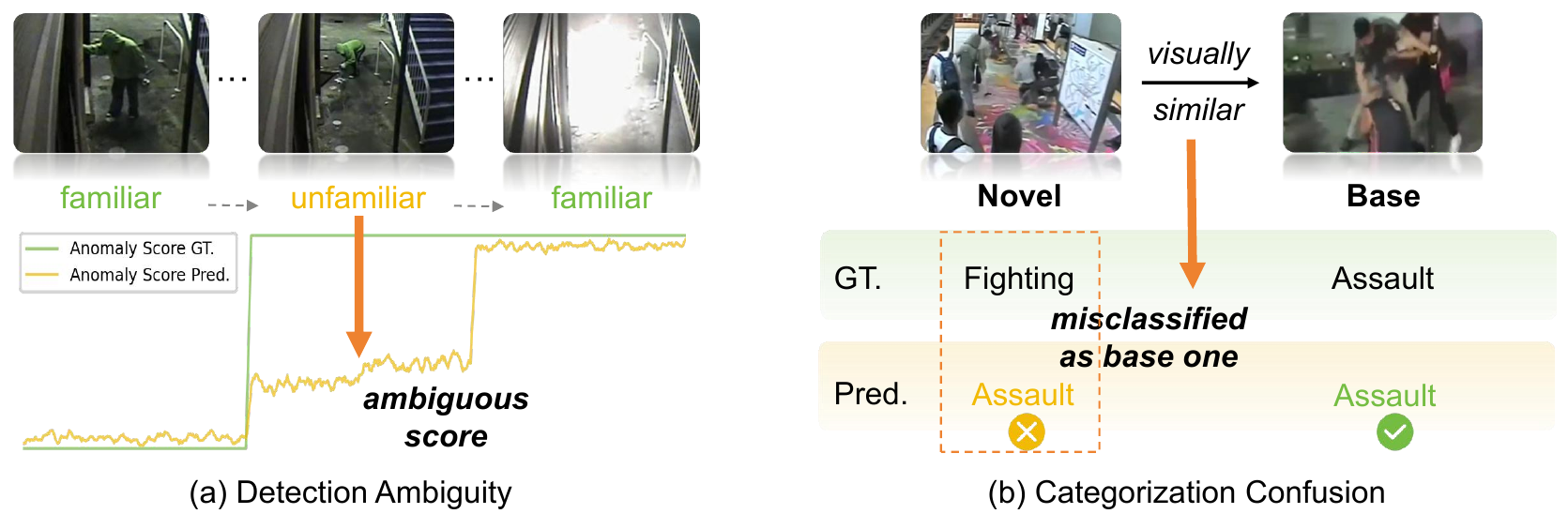}
	\vspace{-5pt}
	\caption{\textbf{Challenges Related to Novel Anomalies.} 
	(a) Detection ambiguity: The model struggles to assign accurate anomaly scores to unfamiliar frames containing novel anomalies. 
	(b) Categorization confusion: Novel anomalies are misclassified as visually similar base instances from the training set.}
	\label{fig:motivation1} 
	\vspace{-5pt}
\end{figure}

Novel anomalies in OVVAD introduce two challenges that remain unexplored by existing methods: (1) \textit{Detection ambiguity}, where the model often lacks sufficient information to accurately assign anomaly scores to unfamiliar data, as shown in \cref{fig:motivation1}(a). Current methods rely on training or fine-tuning the model, which is inherently limited and cannot adapt to the variability of samples in an open setting. (2) \textit{Categorization confusion}, where novel cases visually similar to base cases are misclassified, as shown in \cref{fig:motivation1}(b). OV tasks generally rely on multimodal alignment for categorization. Since the model tends to extract visual features for novel videos similar to base videos, these features are more likely to align with base label encodings, leading to miscategorization. Traditional OV methods use pre-trained encoders to encode text, where the input contains labels with unified templates~\cite{radford2021learning, wu2024building, kim2024retrieval, wu2024open2} or embeddings~\cite{chatterjee2024opening, du2022learning, zhou2022learning}. These methods rely solely on pre-trained encoders without spatial guidance in label encoding, limiting multimodal alignment for novel cases.

\begin{figure} 
	\centering
	\includegraphics[width = 0.9\linewidth]{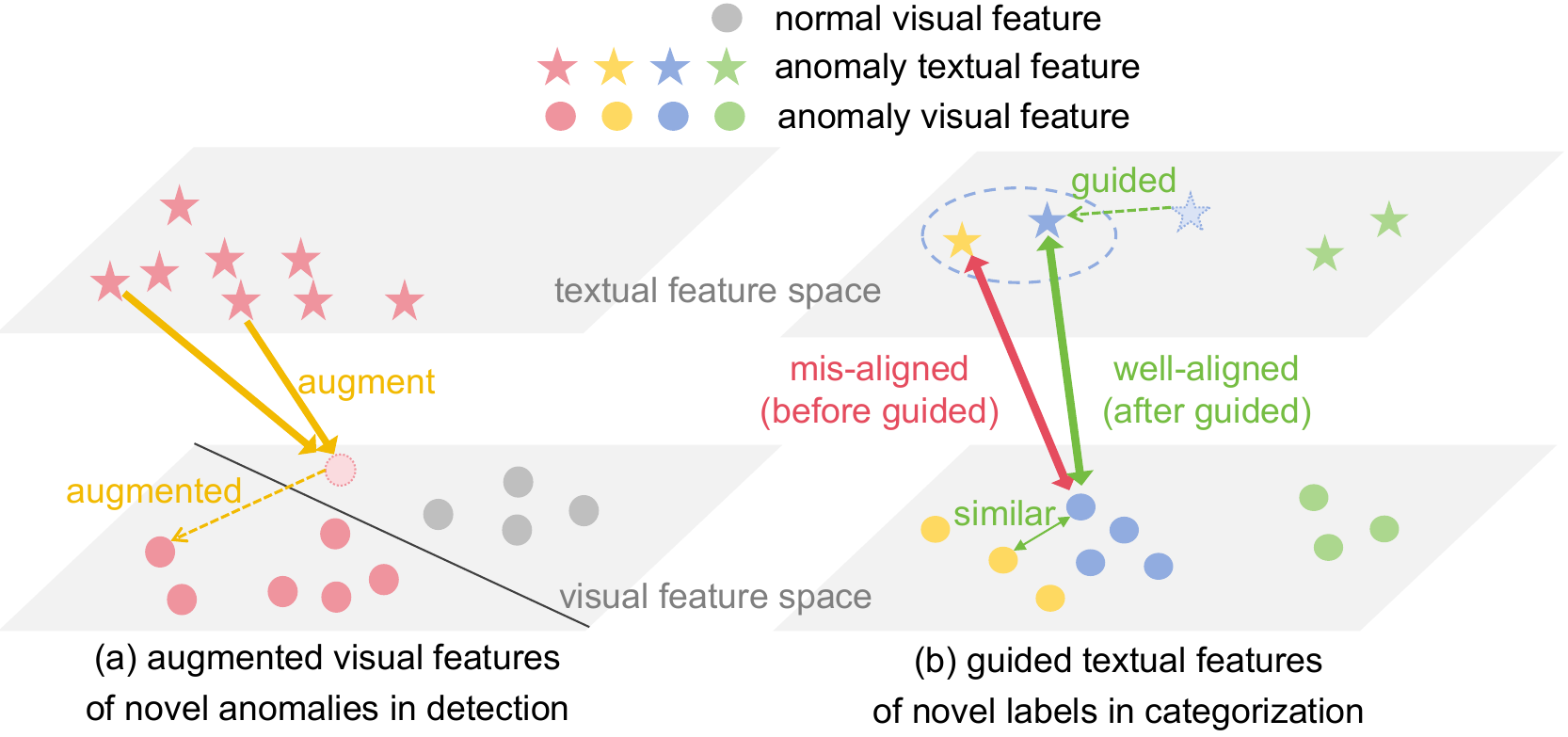} 
	\vspace{-5pt}
	\caption{\textbf{Feature Visualization of Our Design.}
	(a) Text augmentation shifts ambiguous frames to the anomalous feature space. In the static stream, text represents anomaly-related nouns (\textit{e.g.,} ``abandoned fire starter''), while in the dynamic stream, it denotes label descriptions.
	(b) Group-guided text encoding improves the alignment of novel anomalies with novel labels, especially for those resembling base samples.}
	\label{fig:motivation2} 
	\vspace{-5pt}
\end{figure}

To address detection ambiguity, we introduce a \textbf{Text-Augmented Dual Stream} mechanism with dynamic and static streams, each focusing on different visual features augmented by corresponding textual information. The dynamic stream captures sequential information through temporal visual encoding, augmented by label descriptions related to dynamic characteristics. The static stream captures scene information through original contrastive language-image pre-training (CLIP)-encoded visual features, augmented by a concept library related to static characteristics. The complementarity between dynamic and static data is crucial: certain anomalies rely on temporal information, such as tailing, while others depend on contextual cues, such as running on a highway. Synergistic training of the streams ensures mutual supplementation and constraints, delivering comprehensive temporal and contextual information, minimizing overfitting to specific anomaly categories, and improving overall performance. Additionally, the augmentation follows common-sense reasoning. To detect anomalies in real-world scenarios, we first define the anomaly and establish correlations between visual data and anomaly texts, providing a reference for detection within the overall context. Similarly, we augment visual features with relevant anomaly text, providing additional information for detection. As shown in \cref{fig:motivation2}(a), novel visual features that cause ambiguous detections are shifted into the anomalous feature space with support from text, helping the model better assess unfamiliar anomalies.

% \setlength{\textfloatsep}{8pt}
% \begin{figure} 
% \centering
% \begin{subfigure}{0.22\textwidth}
% \centering
% \includegraphics[width	 = 	\linewidth,height	 = 	3.0cm,,keepaspectratio]{sec/images/motivation2_2v4.pdf}
% \caption{visualization in detection}
% \label{fig:motivation2_2} 
% \end{subfigure}
% \begin{subfigure}{0.22\textwidth}
% \centering
% \includegraphics[width	 = 	\linewidth,height	 = 	3.0cm,,keepaspectratio]{sec/images/motivation2_1v4.pdf}
% \caption{visualization in categorization}
% \label{fig:motivation2_1} 
% \end{subfigure}
% \caption{
% \textbf{Feature Visualization of Our Design.}
% (a) Using static stream as an example, where the text consists of nouns describing anomalies. With text-augmented visual data, previously ambiguous frames in novel anomalies become more distinguishable.
% In dynamic stream, the text denote label descriptions. 
% With dual-stream using text-augmented visual data, previously ambiguous novel anomalous data can be more clearly distinguished.
% (b) With the group-guided encoding mechanism, novel anomaly videos similar to base samples achieves more accurate alignment.
% }
% \label{fig:motivation2} 
% \end{figure}

% \setlength{\textfloatsep}{8pt}

To address categorization confusion, we introduce a \textbf{Group-Guided Text Encoding} mechanism, encoding labels using group-based descriptions, with labels sharing similar visual characteristics grouped together. As shown in \cref{fig:motivation2}(b), this mechanism establishes connections between base and novel data through grouping, positioning the encodings of novel labels close to those of base labels, where videos associated with both base and novel labels are visually similar, thereby enhancing multimodal alignment for novel data. For novel labels not grouped with base labels, the descriptions provide contextual support to pre-trained encoders for text encoding, thus enhancing alignment. Compared to previous methods mainly relying on pre-trained models, our approach strengthens the guidance of the feature space for novel labels, achieving more effective alignment for categorization.

% our method transfer knowledge from base cases and enhances the controllability of results for novel cases.

% 根据我们的设计，我们在两个数据集上达到了很好的效果，尤其是提升了难例样本上的表现。

With the \textbf{Anomize} framework, we achieve notable results across both \textsc{XD-Violence}~\cite{wu2020not} and \textsc{UCF-Crime}~\cite{sultani2018real} datasets. For anomaly detection, we obtain a 2.78\% overall improvement on \textsc{XD-Violence} and 8.21\% on novel cases, with \textsc{UCF-Crime} results comparable to a more complex state-of-the-art model. For categorization, we achieve a 25.61\% overall increase in Top-1 accuracy on \textsc{XD-Violence} and 5.71\% on \textsc{UCF-Crime}, with improvements of 56.53\% and 4.49\% on novel cases, respectively.

In summary, our contributions are threefold:

\begin{itemize}
% Insightful Discovery.
% 据我们所知，我们是首个在开放词汇视频异常检测中探索难例样本影响的研究。我们的工作为该领域的未来发展和拓展提供了新的视角与启示。
% \item As we know, we are the first to explore the impact of hard samples in open-vocabulary video anomaly detection. Our findings offer new insights for the future development and expansion of this research field.
% We emphasize the importance of hard samples in OVVAD and find that text guidance and visual augmentation can significantly enhance performance. To address OVVAD, we propose the \textbf{SDM-TG} architecture. 
% By focusing on hard samples in OVVAD, we discover the effectiveness of text guidance and visual augmentation, leading to the \textbf{SDS-TG} framework.
% We discover that guided text encoding and visual augmentation effectively tackle challenges of novel anomalies, 
% Unlike previous methods treating base and novel data equally, our approach offers new insights for OVVAD.
% 异常分数分支上的设计.Innovation Framework.
% 我们提出了一种双检测模块：动态分支利用时序建模后的视觉编码进行预测，静态分支基于原始的视觉编码，两者相互补充与制约。此外，我们引入文本增强模块，通过标签描述文本和名词概念库弥补视觉信息的不足，降低异常判断的模糊性。
% \item We propose a dual detection module. The dynamic module predicts based on temporal visual encodings, while the static module predicts based on original visual encodings. The two modules complement and constrain each other. Additionally, we introduce text augmentation to reduce ambiguity in anomaly detection, utilizing label descriptions and a noun concept library to compensate for limitations in visual information.

	\item To address detection ambiguity, we discover the importance of providing sufficient informational support. We combine dynamic and static streams to effectively constrain and complement each other. Operating at different levels of visual features, each stream is augmented with corresponding textual information, offering comprehensive support for detection.

% To tackle detection ambiguity, we find that providing sufficient informational support makes anomaly visual features more distinguishable. Building on this insight, we introduce a dual-stream mechanism that leverages both dynamic and static visual streams, each augmented with corresponding textual information, to provide comprehensive support for anomaly detection.
% that combines text-augmented visual information at multiple temporal stages.
% by effectively constraining and complementing each other. 
% across various levels of visual features.
% 异常标签分支上的设计.Innovation Framework.
% 我们提出了一种基于大语言模型的标签文本编码引导机制，以增强视觉与文本的对齐效果。通过根据异常视频的视觉相似性对标签进行分组，并使用定制的提示词生成描述文本，该机制确保文本编码与视觉相似性保持一致，从而提升视觉与文本表示之间的对应关系。
% \item We propose a label text encoding guidance mechanism based on large language models to enhance visual-text alignment. We group anomaly labels according to the visual similarity of abnormal videos and generate descriptive texts with tailored prompts to ensure that textual encodings align with visual proximity, thereby improving categorization performance.
% \item To address label confusion in challenging anomaly categorization, we propose Large Language Model-Text Guidance mechanism. Labels are grouped based on the visual similarity of anomalous videos, and the LLM generates descriptive texts with shared elements across groups. This approach ensures that text encodings match visual similarities, improving the alignment between the two modalities.

	\item To tackle categorization confusion, we emphasize the importance of establishing connections between labels to guide their encodings. We propose a text encoding mechanism that groups labels based on visual characteristics and generates corresponding descriptions for encodings.

% To tackle categorization confusion, we discover that establishing semantic connections between labels significantly improves alignment accuracy. Building on this insight, we propose a novel mechanism that organizes labels based on visual characteristics and generating descriptions to guide their encodings.
% This transfers knowledge from base labels for controllable text encoding and improved multimodal alignment.

	\item Our \textbf{Anomize} framework targets the challenges of novel anomalies that remain unexplored, offering new insights for OVVAD and demonstrating superior performance on two widely-used datasets, particularly for novel cases.

\end{itemize}

\section{Related Work}
\label{sec:relatedwork}

\subsection{Video Anomaly Detection}

\paragraph{Semi-Supervised VAD.} 
Existing semi-supervised video anomaly detection (VAD) methods are typically categorized into three groups: one-class classification (OCC), reconstruction-based models, and prediction-based models, all of which are trained exclusively on normal data. OCC models~\cite{zaheer2020old, sabokrou2018adversarially, scholkopf1999support, wu2019deep, wang2019gods} classify anomalies by identifying data points that fall outside a learned hypersphere of normal data. However, defining normality can be ambiguous, which often reduces their effectiveness~\cite{liu2024pixel,xie2023striking}. Reconstruction-based methods~\cite{cong2011sparse, lu2013abnormal, ristea2024self, yang2023video, zhong2022cascade} use deep auto-encoders (DAEs) to learn normal patterns, detecting anomalies through high reconstruction errors. However, DAEs may still reconstruct anomalous frames with low error, weakening detection performance. Prediction models~\cite{liu2018future, liu2021hybrid, hao2022spatiotemporal, feng2021convolutional, lu2019future, lee2019bman}, often utilizing GANs, forecast future frames and identify anomalies by comparing predicted frames with actual frames.

\vspace{-10pt}
\paragraph{Weakly-Supervised VAD.} 
Weakly-supervised video anomaly detection (WSVAD) identifies anomalies using only video-level labels without precise temporal or spatial information. WSVAD methods typically frame the task as a multiple instance learning (MIL) problem~\cite{sultani2018real, zaheer2020claws, li2022self, tian2021weakly, wu2021learning}, where videos are divided into segments, and predictions are aggregated into video-level anomaly scores. Sultani \etal~\cite{sultani2018real} first define the WSVAD paradigm using a deep multiple-instance ranking framework. Recent methods focus on optimizing models. Tian \etal~\cite{tian2021weakly} introduce RTFM, which combines dilated convolutions and self-attention to detect subtle anomalies, while Zaheer \etal~\cite{zaheer2020claws} add a clustering-based normalcy suppression mechanism. Other approaches~\cite{joo2023clip, lv2023unbiased} leverage pre-trained models to gain task-agnostic knowledge. Wu \etal~\cite{wu2024vadclip} propose VadCLIP, which uses CLIP~\cite{radford2021learning} for dual-branch outputs of anomaly scores and labels.

% 对openset-VAD的介绍 
% Open-set VAD aims to train the model based on normality and seen anomalies, and attempts to detect unseen anomalies. This aligns with the requirements of an open-world setting, where systems must handle previously unencountered events effectively.
% openset-VAD的三个例子
% Acsintoae \etal~\cite{acsintoae2022ubnormal} were the first to propose the OV-VAD task, introducing a benchmark dataset and evaluation framework for this task.
% Zhu \etal~\cite{zhu2022towards} integrated evidential deep learning and normalizing flows within a multiple instance learning framework to effectively detect both known and unseen anomalies.
% Hirschorn \etal~\cite{hirschorn2023normalizing} propose a lightweight normalizing flows framework based on human pose graph structures.

\vspace{-10pt}
\paragraph{Open-Set VAD.} 
Open-set VAD models are trained on normal behaviors and base anomalies to detect all anomalies, addressing the challenges of open-world environments. Acsintoae \etal~\cite{acsintoae2022ubnormal} first introduce open-set VAD, along with a benchmark dataset and evaluation framework. Zhu \etal~\cite{zhu2022towards} combine evidential deep learning and normalizing flows within a multiple instance learning framework. Hirschorn \etal~\cite{hirschorn2023normalizing} propose a lightweight normalizing flows framework that utilizes human pose graph structures.

% 对我们方法的介绍
% 在我们的方法中，遵循开放集视频异常检测（open-set VAD）的设置，基于视频级别的异常标签，通过双分支实现异常分数和异常标签输出，可以检测和分类训练集中没有出现的异常。

Our method provides both detection and categorization results in an open setting, focusing on addressing the challenges related to novel anomalies.

\subsection{Open Vocabulary Learning} 
Recent advancements in pre-trained vision-language models~\cite{radford2021learning, jia2021scaling} have spurred significant interest in open vocabulary tasks, including object detection~\cite{gu2021open, zareian2021open, kim2023region}, semantic segmentation~\cite{liang2023open, xu2023side, han2023global}, and action recognition~\cite{wang2021actionclip, lin2024rethinking, ju2022prompting, jia2023generating}.
These studies leverage the pre-trained knowledge of multimodal models, demonstrating strong generalization.
Wu \etal~\cite{wu2024open} first introduce open vocabulary video anomaly detection (OVVAD) using the pre-trained model CLIP. However, most methods emphasize the visual encoder while neglecting the text encoder, limiting zero-shot capabilities. Our method explores the text encoder and incorporates a guided encoding mechanism to enhance multimodal alignment in OVVAD.

\section{Proposed Anomize Method}

\subsection{Overview}

Following Wu \etal~\cite{wu2024open}, we define the training sample set as $D = \{(v_i, y_i)\}_{i = 1}^{N+A}$, which consists of $N$ normal samples $D_n$ and $A$ abnormal samples $D_a$. Here, $v_i$ represents video samples, and $y_i \in \mathbb{C}_{\mathrm{base}}$ denotes the corresponding anomaly labels. Each $v_i \in D_a$ contains at least one anomalous frame, while $v_i \in D_n$ consists entirely of normal frames. The complete label set $\mathbb{C}$ includes both base and novel anomaly labels. The objective of OVVAD is to train a model on $D$ to predict frame-level anomaly scores and video-level anomaly labels from $\mathbb{C}$.

\begin{figure*} 
	\centering
	\includegraphics[width = 0.9\linewidth]{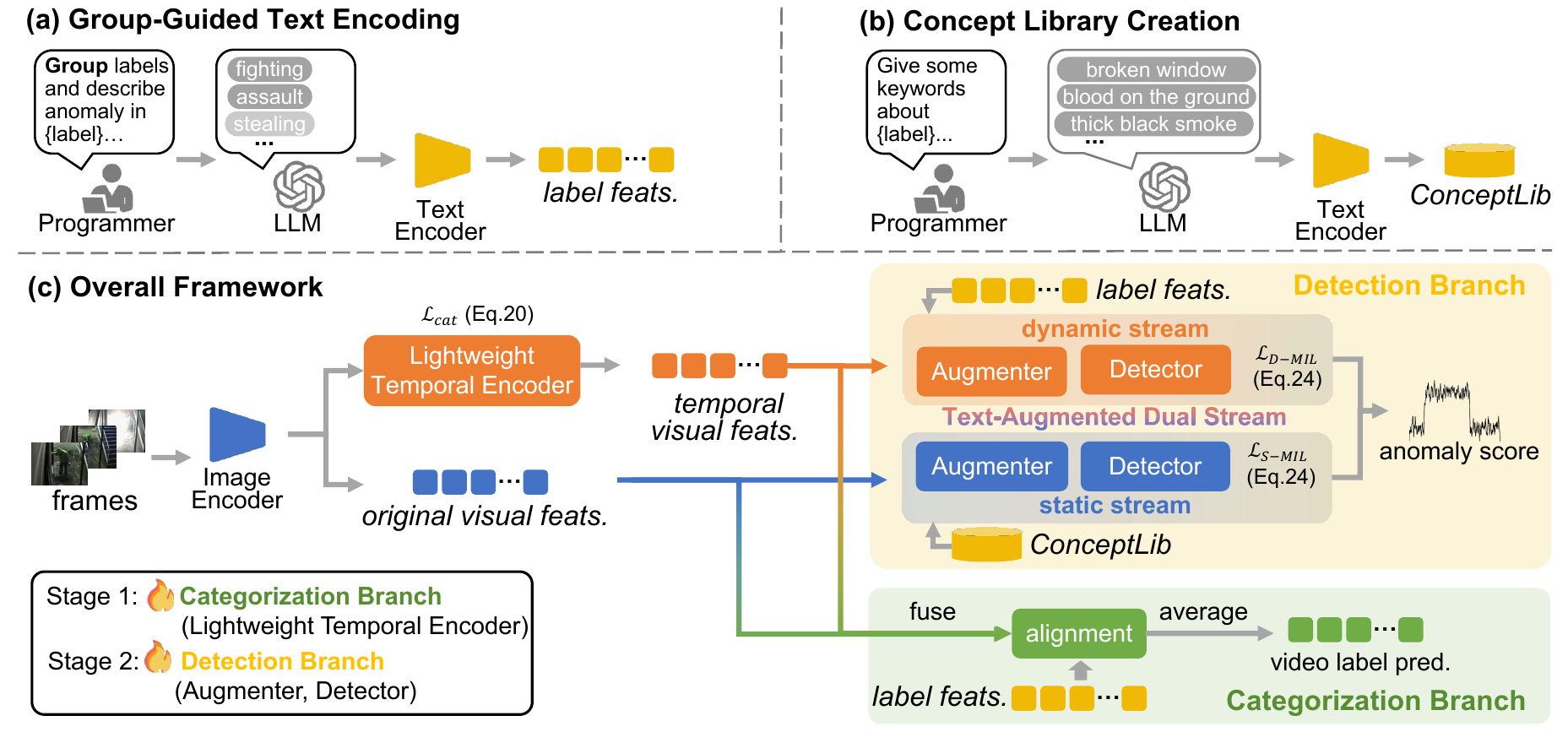}
	\vspace{-5pt}
	\caption{\textbf{Overview of Our Anomize Framework.} (a) Process for obtaining label features via the Group-Guided Text Encoding mechanism. (b) Creation of the concept library $\mathrm{ConceptLib}$ for anomaly detection. (c) The framework processes anomaly labels and video frames to generate frame-level anomaly scores and detected labels. Scoring is performed using a Text-Augmented Dual Stream mechanism, where each stream receives corresponding text and visual features, and the fused scores are produced as output. For labeling, the model aligns label features from the Group-Guided Text Encoding mechanism with the fused original and temporal visual encodings. Both the text and image encoders, pre-trained on CLIP, remain frozen without further optimization.}
	\label{fig:framework} 
	\vspace{-5pt}
\end{figure*}

\cref{fig:framework} illustrates the overview of framework. We leverage the encoder of the pre-trained CLIP model for its strong generalization capabilities. Video frames are processed by the CLIP image encoder $\Phi_{\mathrm{visual}}$ to extract original visual features $x_f \in \mathbb{R}^{n \times d}$, where $n$ is the number of frames and $d$ is the feature dimension. These features are temporally modeled by a lightweight temporal encoder. The original features, augmented by a concept library $\mathrm{ConceptLib}$, pass through the static stream, while temporal features, augmented by label descriptions, pass through the dynamic stream. The prediction from each stream is obtained and aggregated to generate the final frame-level anomaly score $s \in \mathbb{R}^{n \times 1}$. For categorization, a multimodal alignment method is used. A fused visual feature is first generated, and the CLIP text encoder $\Phi_{\mathrm{text}}$ extracts textual features via the group-guided text encoding mechanism. Frame-level predictions are then obtained through alignment and aggregated for the final video-level result $p_{\mathrm{video}}$.

% We first generate a fused visual feature, then use the CLIP text encoder $\Phi_{{text}}$ to generate textual features through the LLM-TG mechanism. 
% Finally, we align the fused features with the anomaly textual features to produce the video-level category prediction.

\subsection{Lightweight Temporal Encoder}
\label{sec:LTE}

% 解释为什么用CLIP视觉编码器
% In this study, we use the frozen $\Phi_{{visual}}$ for visual features, leveraging its pre-trained knowledge for zero-shot classification.

We utilize the frozen $\Phi_{\mathrm{visual}}$ for visual features to leverage its zero-shot capabilities. However, since CLIP is pre-trained on image-text pairs, it lacks temporal modeling for video. Recent methods~\cite{weng2023open, ju2022prompting, wasim2024videogrounding} commonly introduce a temporal encoder. However, this often leads to performance degradation on novel cases, as the additional parameters in the encoder may become specialized for the training set, leading to overfitting. Therefore, we adopt a lightweight long short-term memory (LSTM)~\cite{hochreiter1997long} for temporal modeling, resulting in the temporal visual feature $x_{\mathrm{tem}} \in \mathbb{R}^{n \times d}$:
\begin{equation}
	x_{\mathrm{tem}} = \mathrm{LSTM} \left(x_f \right).
\label{eq:videoEncoding}
\end{equation}

% where ${x}_{{tem}} \in \mathbb{R}^{n \times d}$ is the temporally modeled visual feature.

% 简单介绍一下LSTM
% LSTM is a type of recurrent neural network that captures long-term dependencies through memory cells and gating mechanisms. 
% The forget, input, and output gates regulate information flow, enabling the model to retain relevant temporal patterns while discarding irrelevant ones.
% 解释LSTM不是唯一的选择，记得这里之后要添加实验一节的跳转
% Notably, LSTM is not the only option for lightweight temporal modeling. Other models with relatively few parameters can also be considered. We also experimented with the Transformer architecture and found that it can achieve notable classification performance. The detailed results are presented in the experiment section.
% LSTM, a recurrent neural network, captures long-term dependencies via memory cells and gating mechanisms. 

Other parameter-efficient models may also be suitable, as discussed in the supplementary material.

% We also explored the Transformer architecture, which demonstrated notable classification performance. Detailed results are provided in the experiments section.

\subsection{Group-Guided Text Encoding}
\label{sec:LLM-TG}

Previous methods mainly rely on the generalization capabilities of pre-trained models without task-specific guidance, often leading to categorization confusion. We introduce a group-guided text encoding mechanism to address this. 

% enhancing pre-trained knowledge adaptation and output controllability.
% which enables better adaptation of pre-trained knowledge and improves output controllability. 

We leverage large language models (LLM), specifically GPT-4~\cite{achiam2023gpt}, for textual encoding. We first use the prompt $\mathrm{prompt}_{\mathrm{group}}$ to group labels, ensuring that corresponding videos in each group exhibit high visual similarity. Then, we apply the prompt $\mathrm{prompt}_{\mathrm{desc}}$ to generate text descriptions for each label based on the grouping. These descriptions capture shared elements while emphasizing unique characteristics within each group, ensuring the encodings remain similar yet distinguishable:
\begin{equation}
	\mathrm{result}_{\mathrm{group}} = \mathrm{GPT} \left(\mathrm{prompt}_{\mathrm{group}}, \mathbb{C} \right),
\label{eq:groupEquation}
\end{equation}
\begin{equation}
	\mathrm{result}_{\mathrm{desc}} = \mathrm{GPT} \left(\mathrm{prompt}_{\mathrm{desc}}, \mathrm{result}_{\mathrm{group}} \right),
\label{eq:descriptionEquation}
\end{equation}
where $\mathrm{result}_{\mathrm{group}}$ and $\mathrm{result}_{\mathrm{desc}}$ represent the label groups and their descriptions. The descriptions are then passed into the frozen CLIP text encoder $\Phi_{\mathrm{text}}$ to obtain encodings $t_{\mathrm{desc}} \in \mathbb{R}^{c \times d}$, where $c$ is the number of anomaly labels:
\begin{equation}
	t_{\mathrm{desc}} = \Phi_{\mathrm{text}} \left(\mathrm{result}_{\mathrm{desc}} \right).
\label{eq:textEncoding}
\end{equation}
These encodings are used for multimodal alignment. For the visual features, we combine the temporal and original encodings, preserving the knowledge captured by the pre-trained model:
\begin{equation}
	x_{\mathrm{fused}} = x_{\mathrm{tem}} + \alpha x_f,
\label{eq:fusion2}
\end{equation}
where $\alpha$ is a scalar weight. The prediction probabilities for video frames are expressed as:
\begin{equation}
	p_{\mathrm{frame}} = \frac{x_{\mathrm{fused}} t_{\mathrm{desc}}^\top}{\left\|x_{\mathrm{fused}} \right\| \left\|t_{\mathrm{desc}} \right\|},
\label{eq:framePredictionCLIP}
\end{equation}
where $p_{\mathrm{frame}} \in \mathbb{R}^{n \times c}$ represents the probability distribution over $c$ anomaly labels for each frame. To obtain the video-level prediction, we select the top $M$ probabilities for each label and average these top values, where $M$ is the total number of frames divided by 16:
\begin{equation} 
	p_{\mathrm{avg}} = \sigma \left[\frac{1}{M} \sum_{k = 1}^{M} \mathrm{Top} M \left(p_{\mathrm{frame}}[j] \right) \right]_{j = 1}^{c},
\label{eq:average} 
\end{equation}
where $p_{\mathrm{avg}} \in \mathbb{R}^{c}$ is the average probabilities for each label after applying the softmax function $\sigma(\cdot)$. Finally, the video-level prediction $p_{\mathrm{video}}$ is determined as the label with the highest average probability:
\begin{equation}
	p_{\mathrm{video}} = \arg \max_j p_{\mathrm{avg}}^{(j)}.
\label{eq:videoPrediction}
\end{equation}

% \begin{equation} {P}_{{video}}	 = 	\arg\max_{j} \left( \frac{1}{M} \sum_{k = 1}^{M} {P}_{{top}}^{(j)}[k] \right), \label{eq:videoPrediction} 
% \end{equation}
% where ${P}_{{video}}$ is the video-level prediction, determined by the label with the highest average probability.
% ${P}_{{avg}}^{(j)}$.

\subsection{Augmenter}
\label{sec:Augmenter}

% \begin{figure}[htbp] 
% \centering
% \includegraphics[width = 0.95\linewidth]{sec/images/Augmenter.png} 
% \caption{\textbf{Augmenter.} Visual and textual features are refined through an augmenter to generate enhanced visual features with a focus on abnormal details.}
% \label{fig:Augmenter} 
% \end{figure}
% 在我们方法的设计中，动态模块和静态模块都需要使用文本知识来增强视觉编码。因此，我们提出一个通用的增强器，来完成编码的增强，降低模型的复杂度。
% 具体来说，增强器的输入是视觉编码和T个文本编码,通过将其输入到多头注意力层中，其中，视觉编码作为查询，文本编码作为key和value，提取出与视觉编码最相关的文本特征Erefine，作为补充信息。其次，我们用一个投影层线性映射视觉编码Ecoarse，最后将二者concat并用一个线性层来进行降维，获取到最后的输出Eaug。
% In our proposed method, both the dynamic and static modules leverage textual knowledge to enhance visual encodings. To achieve this, we design a unified augmenter that refines the encodings while simultaneously reducing model complexity.

In our method, both streams utilize text to enhance visual encodings via a unified augmenter. The augmenter takes visual encodings $e_{\mathrm{visual}}$ and textual encodings $e_{\mathrm{text}}$ as input. These are processed by a multi-head attention layer $\mathrm{MHA}(\cdot)$, where $e_{\mathrm{visual}}$ acts as the query and $e_{\mathrm{text}}$ acts as the key and value. This operation extracts the most relevant textual features for supplementation, denoted as $e_{\mathrm{refine}}$:
\begin{equation}
	e_{\mathrm{refine}} = \mathrm{MHA} \left(e_{\mathrm{visual}}, e_{\mathrm{text}}, e_{\mathrm{text}} \right).
\label{eq:attention}
\end{equation}
A fully connected layer $\mathrm{FC}(\cdot)$ linearly projects the visual encoding, which is concatenated with the refined textual features and passed through a multi-layer perceptron $\mathrm{MLP}(\cdot)$ for dimensionality reduction. This results in the augmented output $e_{\mathrm{aug}}$:
\begin{equation}
	e_{\mathrm{aug}} = \mathrm{MLP} \left( \left[e_{\mathrm{refine}}; \mathrm{FC} \left(e_{\mathrm{visual}} \right) \right] \right),
\label{eq:concat}
\end{equation}

\subsection{Text-Augmented Dual Stream}
\label{sec:dual-stream}

In open settings, models may struggle to assess unfamiliar anomalies due to limited information. We propose a Text-Augmented Dual Stream mechanism with complementary dynamic and static streams, each augmented by relevant text to provide sufficient support for detection.

% For the dynamic stream, we predict based on temporally modeled feature augmented by label descriptions, as video anomaly detection relies on temporal cues. 

Since video anomaly detection (VAD) relies on temporal cues, we employ a dynamic stream to predict anomaly scores $s_{\mathrm{dyn}} \in \mathbb{R}^{n \times 1}$ based on refined visual features $f_{\mathrm{aug}} \in \mathbb{R}^{n \times d}$, derived from temporal visual features and augmented by label descriptions via the augmenter in \cref{sec:Augmenter}:
\begin{equation}
	f_{\mathrm{aug}} = \mathrm{Augmenter} \left(x_{\mathrm{tem}}, t_{\mathrm{desc}} \right),
\label{eq:fusion1}
\end{equation}
\begin{equation}
	s_{\mathrm{dyn}} = \mathrm{Sigmoid} \left(\mathrm{FC} \left(f_{\mathrm{aug}} + \mathrm{MLP} \left(f_{\mathrm{aug}} \right) \right) \right),
\label{eq:detector1}
\end{equation}
where $\mathrm{Sigmoid}(\cdot)$ converts predictions to [0, 1].

Since the dynamic stream is limited in scene context, we employ a static stream with original visual features augmented by anomaly-relevant concept data.

% Besides, due to challenges from difficult samples and noise in complex scenes, we utilize text as supplementary information to help the model focus on anomaly-relevant data.
% We create a concept library ${ConceptLib}$ within the model, as outlined in Eq.~\ref{eq:nounEquation}. This library contains key features related to anomalies, extracted from their descriptions and generated by the CLIP model:

Specifically, we create a concept library $\mathrm{ConceptLib}$ containing key features related to anomalies. These features are generated by $\Phi_{\mathrm{text}}$ from various nouns describing significant characteristics of the anomalies:
\begin{equation}
	\mathrm{ConceptLib} = \Phi_{\mathrm{text}} \left(\mathrm{GPT}(\mathrm{prompt}_{\mathrm{conc}}, \mathbb{C}) \right),
\label{eq:nounEquation}
\end{equation}
where $\mathrm{prompt}_{\mathrm{conc}}$ is a prompt for relevant nouns. 
We then compute the cosine similarity between the visual feature of frame $i$, $x_f^{(i)}$, and concept features $h \in \mathrm{ConceptLib}$:
\begin{equation}
	\mathrm{sim} \left(x_f^{(i)}, h \right) = \frac{x_f^{(i)} h^\top}{ \left\|x_f^{(i)} \right\| \left\|h \right\|}.
\label{eq:cosine_similarity}
\end{equation}
The top $K$ relevant concept features $h_f^{(i)} \in \mathbb{R}^{K \times d}$ and their scores $s_f^{(i)} \in \mathbb{R}^{K}$ are selected:
\begin{equation}
	h_f^{(i)}, s_f^{(i)} = \mathrm{Top} K \left(\mathrm{sim} \left(x_f^{(i)}, h \right) \right).
\label{eq:topk_selection}
\end{equation}
The selected features are then weighted by their scores and concatenated to form refined textual features for the video:
\begin{equation}
	h_f^{\mathrm{new}} = \left[\sigma \left(s_f^{(i)} \right) \odot h_f^{(i)} \right]_{i = 1}^{n}.
\label{eq:weighted_features}
\end{equation}
Next, the refined features $h_f^{\mathrm{new}} \in \mathbb{R}^{n \times k \times d}$ and the original visual features $x_f$ are passed through the augmenter to generate the augmented encoding $x_{\mathrm{aug}} \in \mathbb{R}^{n \times d}$:
\begin{equation}
	x_{\mathrm{aug}} = \mathrm{Augmenter} \left(x_f, h_f^{\mathrm{new}} \right).
\label{eq:refinement}
\end{equation}
Similar to the dynamic stream, $x_{\mathrm{aug}}$ is fed into a detector for anomaly score prediction $s_{\mathrm{sta}}$ in the static stream:
\begin{equation}
	s_{\mathrm{sta}} = \mathrm{Sigmoid} \left(\mathrm{FC} \left(x_{\mathrm{aug}} + \mathrm{MLP} \left(x_{\mathrm{aug}} \right) \right) \right).
\label{eq:detector2}
\end{equation}
Finally, the outputs of the two streams are aggregated for the overall anomaly score prediction $s$:
\begin{equation}
	s = \beta s_{\mathrm{dyn}} + \left(1 - \beta \right) s_{\mathrm{sta}},
\label{eq:aggregation2}
\end{equation}
where $\beta$ is a tunable parameter balancing the contributions of the dynamic and static streams.

\subsection{Objective Functions}

\paragraph{First Training Stage.}
In the first stage, we focus on video anomaly categorization to train the LSTM while freezing other modules to prevent optimization conflicts. We use cross-entropy loss $\mathcal{L}_{\mathrm{ce}}$ for categorization. To prevent overfitting to normal data due to class imbalance, we add a separation loss $\mathcal{L}_{\mathrm{sep}}$ to enhance the distinction between normal and anomalous predictions. The loss for the first stage is:
\begin{equation}
	\mathcal{L}_{\mathrm{cat}} = \mathcal{L}_{\mathrm{ce}} + \mathcal{L}_{\mathrm{sep}},
\end{equation}
\begin{equation}
	\mathcal{L}_{\mathrm{ce}} = -\frac{1}{N} \sum_{i = 1}^{N} g_i \log \left(p_{\mathrm{avg}, i} \right),
\label{eq:celoss}
\end{equation}
\begin{equation}
	\mathcal{L}_{\mathrm{sep}} = -\frac{1}{N} \sum_{i = 1}^{N} \left| \max(p_{\mathrm{avg}, i}[1:]) - p_{\mathrm{avg}, i}[0] \right| + 1,
\label{eq:seploss}
\end{equation}
where $p_{\mathrm{avg}, i}$ denotes the predicted probabilities for the $i$-th video, and the normal label is at the first index. $g_i$ is the one-hot ground truth, and $N$ denotes the batch size.

\vspace{-10pt}
\paragraph{Second Training Stage.}
In the second stage, we focus on anomaly detection to train the static and dynamic streams while freezing other modules. Following Wu \etal~\cite{wu2024vadclip}, we apply the MIL loss. Specifically, we first average the top $M$ frame anomaly scores to obtain the video-level prediction $\hat{q}_i$, then compute the loss $\mathcal{L}_{\mathrm{X-MIL}}$ for each stream using binary cross-entropy to quantify the difference between predictions and binary labels $q_i$, where $q_i = 1$ denotes an anomaly and $X \in \{D, S\}$ denotes the type of stream.
% Besides, in this phase, we apply a weight $w_i$ for detecting abnormal videos to address data imbalance, increasing the penalty for incorrect anomaly scores. The loss function is defined as follows:
Additionally, we apply a weight $w_i$ in this phase to tackle data imbalance by increasing the penalty for incorrect scores related to anomalous videos. The loss is defined as follows:
\begin{equation}
	\mathcal{L}_{\mathrm{det}} = \mathcal{L}_{\mathrm{D-MIL}} + \mathcal{L}_{\mathrm{S-MIL}},
\end{equation}
\begin{equation}
	\mathcal{L}_{\mathrm{X-MIL}} = -\frac{1}{N} \sum_{i = 1}^{N} w_i \left( q_i \log \left(\hat{q}_i \right) + \left(1 - q_i \right) \log \left(1 - \hat{q}_i \right) \right).
\label{eq:MILloss}
\end{equation}
% Here, $X \in \{D, S\}$ indicates the different type of stream.
% The weight $w_i$ is assigned to each anomaly.

\section{Experimental Results}
\label{sec:experiments}
% The test set contains 500 violent and 300 non-violent videos. 

\subsection{Datasets and Implementation Details}
\paragraph{Datasets.} We evaluate the performance of Anomize on two widely-used benchmark datasets. \textsc{XD-Violence}~\cite{wu2020not} is the largest dataset focused on violent events in videos, containing 3,954 training videos and 800 testing videos. The videos are collected from movies and YouTube, capturing six types of anomalous events across diverse scenarios. \textsc{UCF-Crime}~\cite{sultani2018real} is a large-scale dataset with 1,610 untrimmed surveillance videos for training and 290 for testing, totaling 128 hours. This dataset includes 13 types of anomalous events, spanning both indoor and outdoor settings and providing broad coverage of real-world scenarios.

% \noindent \textbf{Evaluation metrics.} OVVAD focuses on both the detection and categorization of anomalies. For anomaly detection, we employ standard metrics from previous works~\cite{sultani2018real,wu2020not}. Specifically, for \textsc{UCF-Crime}, we compute the area under the frame-level receiver operating characteristic curve (AUC), capturing the trade-off between true positive and false positive rates. For \textsc{XD-Violence}, we report the area under the frame-level precision-recall curve (AP), reflecting the balance between precision and recall. 
% For anomaly categorization, we report Top-1 accuracy on anomalous test videos from both datasets. During testing, we provide these metrics for all categories combined, in addition to separately for base categories and novel categories.

\vspace{-10pt}
\paragraph{Evaluation Metrics.} For anomaly detection, we employ standard metrics from previous works~\cite{sultani2018real, wu2020not}. Specifically, for \textsc{UCF-Crime}, we compute AUC, which captures the trade-off between true positive and false positive rates. For \textsc{XD-Violence}, we report AP, reflecting the balance between precision and recall. For anomaly categorization, we report Top-1 accuracy on anomalous test videos from both datasets. These metrics are provided for all categories combined, as well as separately for base and novel categories, denoted by the subscripts b and n, respectively.

\vspace{-10pt}
\paragraph{Implementation Details.} 
We implement our model in PyTorch and train it on an RTX 4090 with a 256-frame limit. Using the AdamW optimizer~\cite{kingma2014adam} with a learning rate of $2 \times 10^{-5}$ and a batch size of 32, we train for 16 and 64 epochs in two phases. We use the pre-trained CLIP (ViT-B/16) model. The MLP module contains 2 fully connected layers with GeLU activation. The fusion weight $\alpha$ is 1 during training and 2 during testing. $K$ is 25 for \textsc{XD-Violence} and 5 for \textsc{UCF-Crime}. 
Score weight $\beta$ is 1 on \textsc{XD-Violence} (0 for base categories) and 0.5 on \textsc{UCF-Crime} (0 for novel categories).
% The score weight $\beta$ defaults to 1 in \textsc{XD-Violence} but is reduced to 0 for base categories. In \textsc{UCF-Crime}, $\beta$ defaults to 0.5 but is reduced to 0 for novel categories. 
Loss weight $w_i$ follows the normal-to-anomaly ratio per iteration.

\begin{table}
	\centering
	\footnotesize
	\setlength{\tabcolsep}{4.5pt}
	\begin{tabular}{lccc|ccc}
	\toprule[1.1pt]
	{Method} & {AP} & {AP$_b$} & {AP$_n$} & {AUC} & {AUC$_b$} & {AUC$_n$} \\
	\midrule
	Zhu et al.$\ast$~\cite{zhu2022towards} & 64.40 & - & - & 78.82 & - & -\\
	Sultani et al.~\cite{sultani2018real} & 52.26 & 51.25 & 54.64 & 78.25 & 86.31 & 80.12\\
	Wu et al.~\cite{wu2020not} & 55.43 & 52.94 & 64.10 & 82.24 & 90.62 & 84.13\\
	RTFM~\cite{tian2021weakly} & 58.99 & 55.72 & 65.97 & 84.47 & 92.54 & 85.87\\
	Wu et al.~\cite{wu2024open} & 66.53 & 57.10 & 76.03 & \textbf{86.40} & \textbf{93.80} & \textbf{88.20}\\
	\rowcolor{gray!20}
	{Ours} & \textbf{69.31} & \textbf{57.37} & \textbf{84.24} & \underline{84.49} & \underline{93.00} & \underline{87.05}\\
	\bottomrule[1.1pt]
	\end{tabular}
	\vspace{-5pt}
	\caption{\textbf{Detection Metrics (\%) Comparisons for \textsc{XD-Violence} (left) and \textsc{UCF-Crime} (right).} The best results are highlighted in bold, our method is shaded in gray, the symbol $\ast$ indicates different category divisions, and underlined values represent the second-best results.}
	\vspace{-5pt}
	\label{table:ap_comparison}
\end{table}

\begin{table}
	\centering
	\footnotesize
	\setlength{\tabcolsep}{5pt}
	\begin{tabular}{lccc|ccc}
	\toprule[1.1pt]
	{Method} & {ACC} & {ACC$_b$} & {ACC$_n$} & {ACC} & {ACC$_b$} & {ACC$_n$}\\
	\midrule
	Wu et al.~\cite{wu2024open} & 64.68 & 89.31 & 30.90 & 41.43 & 49.02 & 37.08 \\
	\rowcolor{gray!20}
	{Ours} & \textbf{90.29} & \textbf{92.37} & \textbf{87.43} & \textbf{47.14} & \textbf{56.86} & \textbf{41.57} \\
	\bottomrule[1.1pt]
	\end{tabular}
	\vspace{-5pt}
	\caption{\textbf{Top-1 Accuracy (\%) Comparisons on \textsc{XD-Violence} (left) and \textsc{UCF-Crime} (right).}}
	% The best results are highlighted in bold, and our method is shaded in gray.}
	\vspace{-5pt}
	\label{table:top1acc_comparison}
\end{table}

\subsection{Comparison with State-of-the-Art Methods}

In \cref{table:ap_comparison}, we compare the performance of our method with prior VAD methods, ensuring that all methods use the same visual features from CLIP and adopt an open-set setting. On \textsc{XD-Violence}, we achieve the best performance, with an increase of 2.78\% overall and 8.21\% on novel cases, demonstrating the effectiveness of our method in reducing detection ambiguity. On \textsc{UCF-Crime}, our method achieves competitive results, likely due to the lightweight temporal encoder and segmented training, which limits optimization for the detection branch. Since other studies focus solely on traditional VAD without categorization, we compare our method's categorization performance with a leading study~\cite{wu2024open} in \cref{table:top1acc_comparison}. Our method shows significant improvements, with a 25.61\% gain on \textsc{XD-Violence} and 5.71\% on \textsc{UCF-Crime}, as well as further improvements of 56.53\% and 4.49\% on novel cases, highlighting its effectiveness in reducing categorization confusion.

%Furthermore, since other studies focus only on traditional VAD without categorization, we compare the categorization performance of our method with a leading study~\cite{wu2024open}, as shown in \cref{table:top1acc_comparison}. Our method shows significant improvements, with a 25.61\% gain on \textsc{XD-Violence} and 5.71\% on \textsc{UCF-Crime}. On novel cases, we further improve by 56.53\% and 4.49\%, highlighting the effectiveness of our method in addressing categorization confusion.

\begin{table*}
	\centering
	\footnotesize
	\setlength{\tabcolsep}{8pt}
	\begin{tabular}{cccccccccc|ccc}
	\toprule[1.1pt]
	\multicolumn{3}{c}{{$M_\mathrm{sta}$}} & \multicolumn{3}{c}{{$M_\mathrm{dyn}$}} & \multirow{2}[2]{*}{ST} & \multicolumn{3}{c}{\textsc{XD-Violence}} & \multicolumn{3}{c}{\textsc{UCF-Crime}} \\
	% \multirow{2}[2]{*}{\textbf{AP (\%)}} & 
	% \multirow{2}[2]{*}{\textbf{AP$_b$ (\%)}} & 
	% \multirow{2}[2]{*}{\textbf{AP$_n$ (\%)}} & 
	% \multirow{2}[2]{*}{\textbf{AUC (\%)}} & 
	% \multirow{2}[2]{*}{\textbf{AUC$_b$ (\%)}} & 
	% \multirow{2}[2]{*}{\textbf{AUC$_n$ (\%)}} \\
	\cmidrule(lr){1-3} \cmidrule(lr){4-6} \cmidrule(lr){8-10} \cmidrule(lr){11-13}
	visual & text & $w_i$ & visual & text & $w_i$ & & AP (\%) & AP$_b$ (\%) & AP$_n$ (\%) & AUC (\%) & AUC$_b$ (\%) & AUC$_n$ (\%) \\
	\midrule
	$\surd$ & $\times$ & $\times$ & $\times$ & $\times$ & $\times$ & $\surd$ & 54.75 & 46.30 & 76.61 & 48.47 & 47.72 & 50.05\\
	$\surd$ & $\surd$ & $\times$ & $\times$ & $\times$ & $\times$ & $\surd$ & 59.65 & 56.41 & 75.29 & 52.21 & 59.18 & 48.45\\
	$\surd$ & $\surd$ & $\surd$ & $\times$ & $\times$ & $\times$ & $\surd$ & 59.26 & 57.20 & 74.28 & 84.26 & 92.06 & 86.86 \\
	$\times$ & $\times$ & $\times$ & $\surd$ & $\times$ & $\times$ & $\surd$ & 47.19 & 36.87 & 69.79 & 26.93 & 17.39 & 25.73\\
	$\times$ & $\times$ & $\times$ & $\surd$ & $\surd$ & $\times$ & $\surd$ & 54.98 & 57.79 & 69.40 & 23.06 & 14.68 & 19.74\\
	$\times$ & $\times$ & $\times$ & $\surd$ & $\surd$ & $\surd$ & $\surd$ & 54.16 & 56.03 & 68.19 & 83.21 & 91.94 & 85.43\\
	$\times$ & $\times$ & $\times$ & $\surd$ & $\times$ & $\surd$ & $\surd$ & 48.55 & 51.15 & 59.93 & 80.97 & 90.71 & 83.63\\
	$\surd$ & $\surd$ & $\times$ & $\surd$ & $\surd$ & $\times$ & $\surd$ & 64.92 & 52.93 & 79.91 & 52.37 & 59.44 & 48.61\\
	% $\surd$ & $\surd$ & $\times$ & $\surd$ & $\surd$ & $\surd$ & $\surd$ & 69.31 & 41.80 & 84.24 & 83.01 & 92.10 & 85.08\\
	% $\surd$ & $\surd$ & $\surd$ & $\surd$ & $\surd$ & $\times$ & $\surd$ & 64.92 & 57.46 & 79.91 & 84.47 & 92.14 & 87.05 \\
	$\surd$ & $\surd$ & $\surd$ & $\surd$ & $\surd$ & $\surd$ & $\times$ & 58.22 & \textbf{58.58} & 72.31 & \textbf{84.52} & 92.62 & 86.52\\
	\rowcolor{gray!20}
	$\surd$ & $\surd$ & $\surd$ & $\surd$ & $\surd$ & $\surd$ & $\surd$ & \textbf{69.31} & 57.37 & \textbf{84.24} & 84.49 & \textbf{93.00} & \textbf{87.05} \\
	\bottomrule[1.1pt]
	\end{tabular}
	\vspace{-5pt}
	\caption{\textbf{Effectiveness of Dynamic $M_{\mathrm{dyn}}$ and Static $M_{\mathrm{sta}}$ Streams with Text-Augmented Visual Data, Additional Loss Weight $w_i$, and Segmented Training (ST) on \textsc{XD-Violence} (left) and \textsc{UCF-Crime} (right).} In $M_{\mathrm{sta}}$, the visual data is the original visual feature output by CLIP, while in $M_{\mathrm{dyn}}$, it is derived from a lightweight temporal encoder.}
	% The best results are highlighted in bold.
	\vspace{-5pt}
	\label{table:xd_AP_ablation}
\end{table*}

\begin{table*}
% \vspace{-10pt}
	\centering
	\footnotesize
	\setlength{\tabcolsep}{9pt}
	\begin{tabular}{cccccccc|ccc}
	\toprule[1.1pt]
	\multirow{2}[2]{*}{{$E_\mathrm{tem}$}} & \multirow{2}[2]{*}{{$M_\mathrm{group}$}} & \multirow{2}[2]{*}{{$F_\mathrm{fus}$}} & \multirow{2}[2]{*}{{$\mathcal{L}_\mathrm{sep}$}} & \multirow{2}[2]{*}{ST} & \multicolumn{3}{c}{\textsc{XD-Violence}} & \multicolumn{3}{c}{\textsc{UCF-Crime}} \\
	\cmidrule(lr){6-8} \cmidrule(lr){9-11} 
	& & & & & {ACC (\%)} & {ACC$_b$ (\%)} & {ACC$_n$ (\%)} & {ACC (\%)} & {ACC$_b$ (\%)} & {ACC$_n$ (\%)}\\ 
	\midrule
	$\times$ & $\times$ & $\times$ & $\times$ & $\surd$ & 45.92 & 74.81 & 6.28 & 35.71 & 56.86 & 23.60\\
	$\surd$ & $\times$ & $\times$ & $\times$ & $\surd$ & 53.42 & 92.37 & 0 & 25.71 & 70.59 & 0\\
	$\surd$ & $\surd$ & $\times$ & $\times$ & $\surd$ & 56.07 & 92.75 & 5.76 & 27.86 & 68.63 & 4.49\\
	$\surd$ & $\times$ & $\surd$ & $\times$ & $\surd$ & 56.51 & \textbf{94.66} & 4.19 & 27.14 & \textbf{74.51} & 0\\
	$\surd$ & $\surd$ & $\surd$ & $\times$ & $\surd$ & 90.07 & 91.98 & 87.43 & 46.43 & 56.86 & 40.45\\
	$\surd$ & $\surd$ & $\surd$ & $\surd$ & $\times$ & 89.95 & 91.98 & 86.91 & 46.43 & 52.94 & \textbf{42.70}\\
	\rowcolor{gray!20}
	$\surd$ & $\surd$ & $\surd$ & $\surd$ & $\surd$ & \textbf{90.29} & 92.37 & \textbf{87.43} & \textbf{47.14} & 56.86 & 41.57 \\
	\bottomrule[1.1pt]
	\end{tabular}
	\vspace{-5pt}
	\caption{\textbf{Effectiveness of the Lightweight Temporal Encoder $E_\mathrm{tem}$, Group-Guided Text Encoding Mechanism $M_\mathrm{group}$, Fusion Function $F_\mathrm{fus}$, Additional Separation Loss $\mathcal{L}_\mathrm{sep}$, and Segmented Training (ST) on \textsc{XD-Violence} (left) and \textsc{UCF-Crime} (right).}} 
	% The best results are highlighted in bold.}
	\vspace{-5pt}
	\label{table:top1_ablation}
\end{table*}

\subsection{Ablation Studies}
\paragraph{Effectiveness of Lightweight Temporal Encoder.} 
The experiments confirm the importance of temporal information in video-level tasks. \cref{table:xd_AP_ablation} shows that the dynamic stream, with temporal encoding, provides useful temporal cues and complements the static stream effectively. While relying solely on temporal information performs poorly due to noise, the dynamic stream becomes more effective with loss weighting and text augmentation. \cref{table:top1_ablation} shows that adding the temporal encoder improves performance on base cases, but without further guidance, the lightweight encoder still introduces confusion for novel anomalies.

\vspace{-10pt}
\paragraph{Effectiveness of Group-Guided Text Encoding Mechanism.} 
Comparisons in \cref{table:top1_ablation} between the second and third rows, or the fourth and fifth rows, on both datasets show that textual encodings based on group descriptions outperform the baseline, demonstrating the importance of our text encoding mechanism.

% Refer to \textcolor{red}{supplementary material} for prompt design details.

\vspace{-10pt}
\paragraph{Effectiveness of Text Augmentation.} 
As shown in \cref{table:xd_AP_ablation}, text augmentation in both dynamic and static streams generally reduces detection ambiguity by compensating for the limitations of visual features. The dynamic stream with only text augmentation shows a slight drop on \textsc{UCF-Crime}, suggesting noise in the temporal encodings. However, when combined with the loss weight, it demonstrates the importance of text, as shown in Rows 6 and 7.

\vspace{-10pt}
\paragraph{Effectiveness of Integrating Dynamic and Static Streams.} 
\cref{table:xd_AP_ablation} shows that integrating the two streams is generally more effective than using them independently, as they complement and constrain each other, which is evident from the comparisons in the third, sixth, and last rows.
% Applying loss weight only to the dynamic stream (Row 9) on \textsc{UCF-Crime} causes a slight drop compared to Row 6, likely due to training imbalances. Jointly training both streams with the same settings consistently outperforms single-stream methods.

\vspace{-10pt}
\paragraph{Effectiveness of Additional Loss Design.} 
\cref{table:xd_AP_ablation} emphasizes the importance of loss weight $w_i$, especially when training dynamic and static streams together. 
% An exception occurs when training dynamic stream alone with loss weight, likely due to limited parameters causing training confusion. However, the advantages of loss weight are evident when both streams are integrated. 
On \textsc{XD-Violence}, adding $w_i$ slightly degrades single-stream training but significantly benefits integrated streams.
Besides, \cref{table:top1_ablation} shows that adding separation loss $\mathcal{L}_\mathrm{sep}$ improves performance, effectively addressing class imbalance.

\vspace{-10pt}
\paragraph{Effectiveness of Segmented Training.} 
Results in \cref{table:xd_AP_ablation} and \cref{table:top1_ablation} show that segmented training generally outperforms single-phase training, especially for novel cases, confirming that single-phase training may cause optimization conflicts and increase overfitting risks. On \textsc{UCF-Crime}, single-phase training shows slightly better categorization performance on novel categories, likely due to randomness caused by optimization conflicts, as reflected in the poor performance on base categories. 

% \noindent Refer to supplementary material for more ablation results.

\begin{figure*}
	\centering
	\includegraphics[width = 0.9 \linewidth]{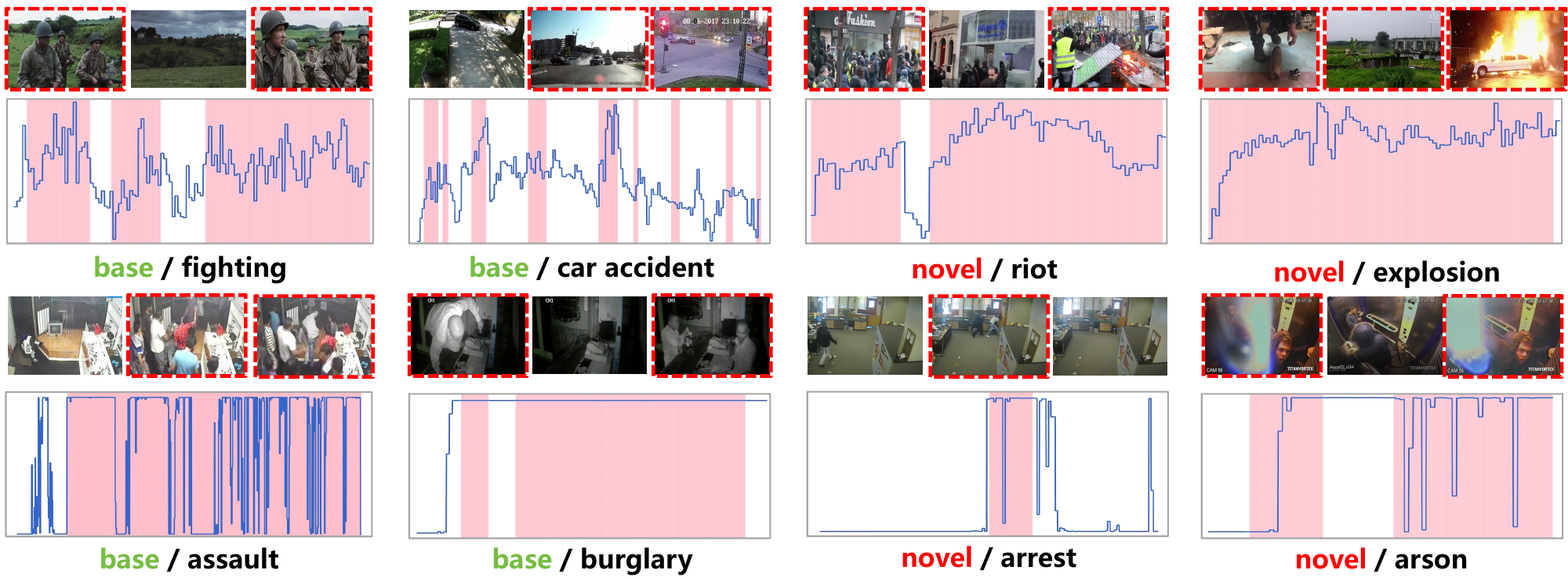}
	\vspace{-5pt}
	\caption{\textbf{Qualitative Results for Anomaly Detection.} The first and second rows present results on \textsc{XD-Violence} and \textsc{UCF-Crime} respectively. Red boxes and rectangles highlight the ground-truth anomalous frames, while blue lines represent predicted anomaly scores.}
	\label{fig:detection_Qualitative_Results}
	\vspace{-5pt}
\end{figure*} 

\subsection{Qualitative results}

\cref{fig:detection_Qualitative_Results} presents qualitative results for anomaly detection, featuring two base and two novel categories from each dataset to cover all label groups. The comparison between the blue predicted score lines and the pink rectangles for ground truth demonstrates the effectiveness of our model in detecting anomalies. Notably, our method demonstrates its capability to handle novel cases with minimal detection ambiguity, highlighting the strong support of the text-augmented dual stream.
% Additionally, the varying performance across different datasets provides insights for the fine-grained optimization of text guidance and visual encoding.

% Notably, our approach exhibits minimal detection ambiguity on hard samples, demonstrating its capability to handle such cases.

\begin{figure*}
	\centering
	\includegraphics[width = 0.9 \linewidth]{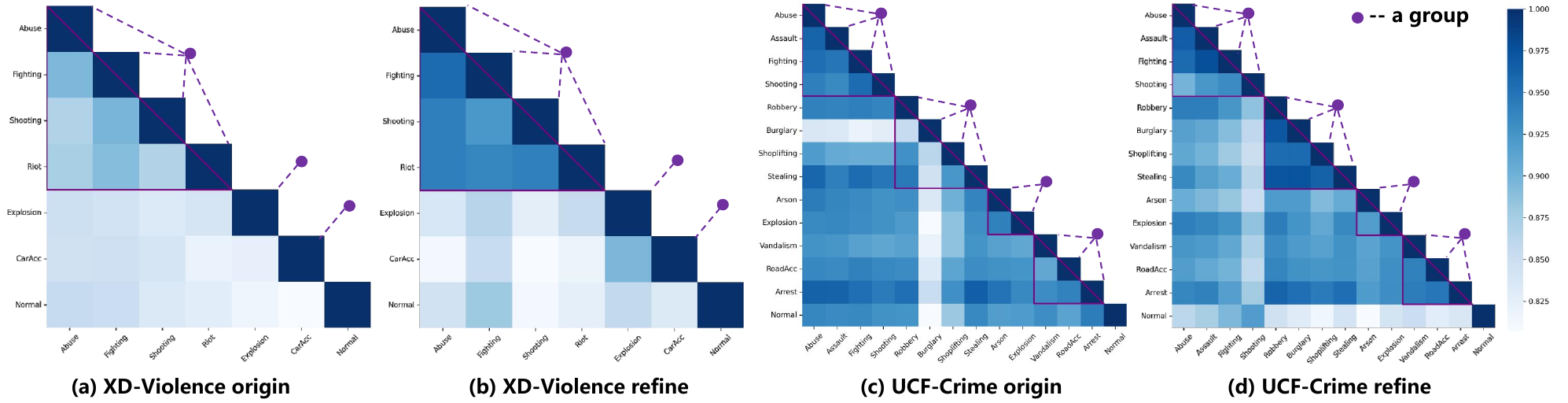}
	\vspace{-5pt}
	\caption{\textbf{Similarity Matrices of Textual Encoding.} (a) and (c) depict results using encodings from the original label data, while (b) and (d) show improvements achieved with the group-guided text encoding mechanism.}
	\label{fig:confusion_matrice}
	\vspace{-5pt}
\end{figure*} 

% \begin{figure*}
% 	\centering
% 	% 第一张图片
% 	% \begin{subfigure}{0.2318\textwidth}
% 	\begin{tabular}{cccc}
% 	\includegraphics[width = 0.2 \linewidth]{sec/images/UCF-origin.pdf} & 
% 	\includegraphics[width = 0.2 \linewidth]{sec/images/UCF-refine.pdf} & 
% 	\includegraphics[width = 0.2 \linewidth]{sec/images/XD-origin.pdf} & 
% 	\includegraphics[width = 0.2 \linewidth]{sec/images/XD-refine.pdf} \\
% 	\small{(a) \textsc{UCF-Crime} origin} & \small{(b) \textsc{UCF-Crime} refine} & \small{(c) \textsc{XD-Violence} origin} & \small{(d) \textsc{XD-Violence} refine} 
% 	\end{tabular}
% 	\caption{\textbf{Similarity Matrices of Anomaly Textual Encoding.} (a) and (c) depict results using encodings from the original label data, while (b) and (d) show improvements achieved with the group-guided text encoding mechanism.}
% 	\label{fig:confusion_matrice}
% 	\vspace{-5pt}
% \end{figure*} 

\cref{fig:confusion_matrice} shows similarity matrices of textual encodings, where labels indicated by the same dot are grouped together. \cref{fig:confusion_matrice}(a) and (c) show results from encoding text using only label text, where visually similar anomalies lack corresponding textual similarity, revealing the limitations of relying solely on pre-trained models. \cref{fig:confusion_matrice}(b) and (d) present results after being guided, where textual similarities within groups are enhanced to better align with visual similarities. This demonstrates the benefit of establishing label connections to guide encoding. 
\begin{table}
	\centering	
	\footnotesize
	\begin{tabular}{l|cc|cc}
	\toprule[1.1pt]
	\multirow{2}{*}{\diagbox[width = 6em]{Train}{Test}} & 
	\multicolumn{2}{c|}{{\textsc{XD-Violence}}} & 
	\multicolumn{2}{c}{{\textsc{UCF-Crime}}} \\
	\cmidrule{2-5}
	& {AP (\%)} & {ACC (\%)} & {AUC (\%)} & {ACC (\%)} \\
	\midrule
	{\textsc{XD-Violence}} & \cellcolor{gray!20}\textbf{69.31} & \cellcolor{gray!20}\textbf{90.29} & 81.69 & 45.71 \\
	{\textsc{UCF-Crime}} & 66.16 & 85.87 & \cellcolor{gray!20}\textbf{84.68} & \cellcolor{gray!20}\textbf{47.14} \\
	\bottomrule[1.1pt]
	\end{tabular}
	\vspace{-5pt}
	\caption{\textbf{Cross-Dataset Detection and Categorization Results.}} 
	% The best results are highlighted in bold.}
	\label{table:crossdataset_performace}
	\vspace{-5pt}
\end{table}

\subsection{Adaptability and Generalization}

% \subsubsection{Analysis of cross-dataset ability}

% \noindent 
% As shown in the \cref{table:crossdataset_performace}, training the model in an open-set manner on one dataset and testing it on another dataset yields comparable results to training directly on the target dataset. This holds true for both anomaly categorization and detection, demonstrating the adaptability of our method and its effectiveness in cross-dataset settings.

\paragraph{Analysis of Cross-Dataset Ability.} 
% \cref{table:crossdataset_performace} shows that training the model in an open-set manner on one dataset and testing it on another yields results comparable to training directly on the target dataset, demonstrating the adaptability and effectiveness of our method in cross-dataset anomaly detection and categorization.
\cref{table:crossdataset_performace} shows that training on one dataset in an open-set manner and testing on another achieves results comparable to direct training on the target dataset, highlighting the adaptability of our method.

% From \cref{table:crossdataset_performace}, the abnormal label categorization branch shows strong adaptability, as a model trained in an open-set manner on one dataset performs well on another dataset, with results close to those obtained through open-set training on the target dataset. In the abnormal score detection branch, the UCF-Crime dataset shows minimal decline, while the XD-Violence dataset experiences a more noticeable drop when tested on the model trained with UCF-Crime data. This may result from the diverse abnormal scenarios in UCF-Crime introducing noise during training, impacting performance.
% \subsubsection{Analysis of open-vocabulary ability}

\begin{table}
% \vspace{-8pt}
	\centering
	\footnotesize
	\setlength{\tabcolsep}{10pt}
	\begin{tabular}{lccc}
	\toprule[1.1pt]
	{Setting} & {ACC (\%)} & {ACC$_b$ (\%)} & {ACC$_n$ (\%)} \\
	\midrule
	\rowcolor{gray!20}
	orig. & 90.29 & \textbf{92.37} & 87.43 \\
	+ shoplifting & \textbf{90.51} & 92.37 & \textbf{88.21} \\
	+ arrest & 89.96 & 92.37 & 86.73 \\
	+ arson & 88.77 & 92.37 & 84.08 \\
	+ assault & 87.28 & 87.40 & 87.11 \\
	+ labels & 85.21 & 86.64 & 83.25 \\
	\midrule
	\rowcolor{gray!20}
	orig. & 47.14 & \textbf{56.86} & 41.57 \\
	+ riot & \textbf{65.69} & 56.86 & \textbf{68.09} \\
	+ labels & 45.00 & 54.90 & 39.33 \\
	\bottomrule[1.1pt]
	\end{tabular}
	\vspace{-5pt}
	\caption{\textbf{Impact of Data Addition on Top-1 Accuracy for \textsc{XD-Violence} (top) and \textsc{UCF-Crime} (bottom).} ``+ shoplifting'' denotes the addition of both videos and labels, while ``+ labels'' indicates adding new labels only.}
	\label{table:afterAddingData_performance}
	\vspace{-5pt}
\end{table}

\vspace{-10pt}
\paragraph{Analysis of Open Vocabulary Ability.} 
The open vocabulary ability is demonstrated by stable categorization performance after adding novel anomaly data, which we further validate by evaluating top-1 accuracy with the added data, as shown in \cref{table:afterAddingData_performance}. The additional data for one dataset is sourced from another. On \textsc{XD-Violence}, incorporating both same-group (\textit{e.g.}, arson) and different-group data (\textit{e.g.}, shoplifting) leads to stable performance, though assault shows the clearest decline due to confusion with similar labels like fighting. On \textsc{UCF-Crime}, incorporating riot leads to notable improvement, with most instances accurately categorized despite being grouped with four other labels. Although the addition of new labels introduces some confusion, the performance remains robust across both datasets.

%This confirms that the group-guided text encoding mechanism enhances generalization.
%$$, allowing the model to excel in open vocabulary tasks and surpass earlier models using only pre-trained knowledge.

% Detailed information can be found in the supplementary materials.
% Refer to \textcolor{red}{supplementary material} for detailed information.
% \begin{table}
%	 \centering
%	 \caption{Impact of Label Addition on Top-1 and Top-5 Accuracy in XD-Violence (top) and UCF-Crime (bottom).}
%	 \resizebox{0.43\textwidth}{!}{ % 调整表格宽度以适应页面
%	 \begin{tabular}{ccccc}
%	 \toprule[1.1pt]
%	 \textbf{setting} & \textbf{orig. top1} & \textbf{+ labels top1} & \textbf{orig. top5} & \textbf{+ labels} \\
%	 \midrule
%	 ACC (\%) & \textbf{90.29} & 85.21 & \textbf{99.56} & 94.70\\
%	 ACC$_b$ (\%) & \textbf{92.37} & 86.64 & \textbf{100} & 95.42 \\
%	 ACC$_n$ (\%) & \textbf{87.43} & 83.25 & \textbf{98.95} & 93.72 \\
%	 \bottomrule[1.1pt]
%	 \end{tabular}
%	 }
%	 \label{table:afterAddingLabel_XD_performance}
% \end{table}

\section{Conclusion}
\label{sec:conclusion}
% 在本论文中，我们介绍了 SDM-TG，这是一种旨在解决开放词汇视频异常检测中具有挑战性样本的标签混淆和分数模糊问题的框架。SDM-TG 采用 LLM-TG 机制，以预训练模型为基础指导文本编码，从而使多模态对齐更加可控并减少标签混淆。通过动态和静态模块使用文本增强器丰富视觉编码中的异常相关细节，该框架提高了对不熟悉样本的检测能力，并解决了分数模糊问题。在 UCF-Crime 和 XD-Violence 数据集上的实验验证了我们方法的有效性。

We propose the Anomize framework to address detection ambiguity and categorization confusion in open vocabulary video anomaly detection (OVVAD). By augmenting visual encodings with anomaly-related text through a dual-stream mechanism, Anomize improves the detection of unfamiliar samples and resolves ambiguity. Additionally, a group-guided text encoding mechanism enhances multimodal alignment and reduces categorization confusion. Experiments on \textsc{XD-Violence} and \textsc{UCF-Crime} demonstrate the effectiveness of our method.
% In the future, we will explore feature-level disentanglement for dynamic and static elements to minimize redundancy in current semantic-level separation.
% \section*{Limitations}
% \paragraph{Limitations.}
% In this paper, we distinguish dynamic and static elements at the semantic level, which may introduce redundancy. In the future, we will explore feature-level disentanglement to further enhance performance.
% \section*{Acknowledgments}
\vspace{-10pt}

\paragraph{Acknowledgments.} 
% This work was supported in part by the National Natural Science Foundation of China under Grants 62171325 and 62271361, and by the Hubei Provincial Key Research and Development Program under Grant 2024BAB039. The numerical calculations in this paper have been done on the supercomputing system in the Supercomputing Center of Wuhan University.
This work was supported by the National Natural Science Foundation of China (Grants 62171325, 62271361), the Hubei Provincial Key Research and Development Program (Grant 2024BAB039), and the supercomputing system at Wuhan University.


% \clearpage
% \setcounter{page}{1}
% \maketitlesupplementary

{
 \small
 \bibliographystyle{ieeenat_fullname}
 \bibliography{Anomize}

\begin{thebibliography}{54}
\providecommand{\natexlab}[1]{#1}
\providecommand{\url}[1]{\texttt{#1}}
\expandafter\ifx\csname urlstyle\endcsname\relax
  \providecommand{\doi}[1]{doi: #1}\else
  \providecommand{\doi}{doi: \begingroup \urlstyle{rm}\Url}\fi

\bibitem[Acsintoae et~al.(2022)Acsintoae, Florescu, Georgescu, Mare, Sumedrea, Ionescu, Khan, and Shah]{acsintoae2022ubnormal}
Andra Acsintoae, Andrei Florescu, Mariana{-}Iuliana Georgescu, Tudor Mare, Paul Sumedrea, Radu~Tudor Ionescu, Fahad~Shahbaz Khan, and Mubarak Shah.
\newblock Ubnormal: New benchmark for supervised open-set video anomaly detection.
\newblock In \emph{Proc. {IEEE/CVF} Conf. Comput. Vis. Pattern Recognit.}, pages 20111--20121, 2022.

\bibitem[Chatterjee et~al.(2023)Chatterjee, Sener, Ma, and Yao]{chatterjee2024opening}
Dibyadip Chatterjee, Fadime Sener, Shugao Ma, and Angela Yao.
\newblock Opening the vocabulary of egocentric actions.
\newblock In \emph{Adv. Neural Inf. Process. Syst.}, 2023.

\bibitem[Cong et~al.(2011)Cong, Yuan, and Liu]{cong2011sparse}
Yang Cong, Junsong Yuan, and Ji Liu.
\newblock Sparse reconstruction cost for abnormal event detection.
\newblock In \emph{Proc. {IEEE/CVF} Conf. Comput. Vis. Pattern Recognit.}, pages 3449--3456, 2011.

\bibitem[Du et~al.(2022)Du, Wei, Zhang, Shi, Gao, and Li]{du2022learning}
Yu Du, Fangyun Wei, Zihe Zhang, Miaojing Shi, Yue Gao, and Guoqi Li.
\newblock Learning to prompt for open-vocabulary object detection with vision-language model.
\newblock In \emph{Proc. {IEEE/CVF} Conf. Comput. Vis. Pattern Recognit.}, pages 14064--14073, 2022.

\bibitem[Feng et~al.(2021)Feng, Song, Chen, Chen, Ni, and Chen]{feng2021convolutional}
Xinyang Feng, Dongjin Song, Yuncong Chen, Zhengzhang Chen, Jingchao Ni, and Haifeng Chen.
\newblock Convolutional transformer based dual discriminator generative adversarial networks for video anomaly detection.
\newblock In \emph{Proc. ACM Int. Conf. Multimedia}, pages 5546--5554, 2021.

\bibitem[Gu et~al.(2022)Gu, Lin, Kuo, and Cui]{gu2021open}
Xiuye Gu, Tsung{-}Yi Lin, Weicheng Kuo, and Yin Cui.
\newblock Open-vocabulary object detection via vision and language knowledge distillation.
\newblock In \emph{Proc. Int. Conf. Learn. Represent.}, 2022.

\bibitem[Han et~al.(2023)Han, Liu, Liew, Ding, Liu, Wang, Tang, Yang, Feng, Zhao, and Wei]{han2023global}
Kunyang Han, Yong Liu, Jun~Hao Liew, Henghui Ding, Jiajun Liu, Yitong Wang, Yansong Tang, Yujiu Yang, Jiashi Feng, Yao Zhao, and Yunchao Wei.
\newblock Global knowledge calibration for fast open-vocabulary segmentation.
\newblock In \emph{Proc. {IEEE/CVF} Int. Conf. Comput. Vis.}, pages 797--807, 2023.

\bibitem[Hao et~al.(2022)Hao, Li, Wang, Wang, and Gao]{hao2022spatiotemporal}
Yi Hao, Jie Li, Nannan Wang, Xiaoyu Wang, and Xinbo Gao.
\newblock Spatiotemporal consistency-enhanced network for video anomaly detection.
\newblock \emph{Pattern Recognit.}, 121:\penalty0 108232, 2022.

\bibitem[Hirschorn and Avidan(2023)]{hirschorn2023normalizing}
Or Hirschorn and Shai Avidan.
\newblock Normalizing flows for human pose anomaly detection.
\newblock In \emph{Proc. {IEEE/CVF} Int. Conf. Comput. Vis.}, pages 13499--13508, 2023.

\bibitem[Hochreiter and Schmidhuber(1997)]{hochreiter1997long}
Sepp Hochreiter and J{\"{u}}rgen Schmidhuber.
\newblock Long short-term memory.
\newblock \emph{Neural Comput.}, 9\penalty0 (8):\penalty0 1735--1780, 1997.

\bibitem[Jia et~al.(2021)Jia, Yang, Xia, Chen, Parekh, Pham, Le, Sung, Li, and Duerig]{jia2021scaling}
Chao Jia, Yinfei Yang, Ye Xia, Yi{-}Ting Chen, Zarana Parekh, Hieu Pham, Quoc~V. Le, Yun{-}Hsuan Sung, Zhen Li, and Tom Duerig.
\newblock Scaling up visual and vision-language representation learning with noisy text supervision.
\newblock In \emph{Proc. Int. Conf. Mach. Learn.}, pages 4904--4916, 2021.

\bibitem[Jia et~al.(2024)Jia, Luo, Chang, Dang, Han, Wang, Dai, Dang, and Wang]{jia2023generating}
Chengyou Jia, Minnan Luo, Xiaojun Chang, Zhuohang Dang, Mingfei Han, Mengmeng Wang, Guang Dai, Sizhe Dang, and Jingdong Wang.
\newblock Generating action-conditioned prompts for open-vocabulary video action recognition.
\newblock In \emph{Proc. {ACM} Int. Conf. Multimedia}, pages 4640--4649, 2024.

\bibitem[Joo et~al.(2023)Joo, Vo, Yamazaki, and Le]{joo2023clip}
Hyekang~Kevin Joo, Khoa Vo, Kashu Yamazaki, and Ngan Le.
\newblock {CLIP-TSA:} clip-assisted temporal self-attention for weakly-supervised video anomaly detection.
\newblock In \emph{Proc. {IEEE} Int. Conf. Image Process.}, pages 3230--3234, 2023.

\bibitem[Ju et~al.(2022)Ju, Han, Zheng, Zhang, and Xie]{ju2022prompting}
Chen Ju, Tengda Han, Kunhao Zheng, Ya Zhang, and Weidi Xie.
\newblock Prompting visual-language models for efficient video understanding.
\newblock In \emph{Proc. Eur. Conf. Comput. Vis.}, pages 105--124, 2022.

\bibitem[Kim et~al.(2023)Kim, Angelova, and Kuo]{kim2023region}
Dahun Kim, Anelia Angelova, and Weicheng Kuo.
\newblock Region-aware pretraining for open-vocabulary object detection with vision transformers.
\newblock In \emph{Proc. {IEEE/CVF} Conf. Comput. Vis. Pattern Recognit.}, pages 11144--11154, 2023.

\bibitem[Kim et~al.(2024)Kim, Cho, Kim, and Kim]{kim2024retrieval}
Jooyeon Kim, Eulrang Cho, Sehyung Kim, and Hyunwoo~J. Kim.
\newblock Retrieval-augmented open-vocabulary object detection.
\newblock In \emph{Proc. {IEEE/CVF} Conf. Comput. Vis. Pattern Recognit.}, pages 17427--17436, 2024.

\bibitem[Kingma and Ba(2015)]{kingma2014adam}
Diederik~P. Kingma and Jimmy Ba.
\newblock Adam: {A} method for stochastic optimization.
\newblock In \emph{Proc. Int. Conf. Learn. Represent.}, 2015.

\bibitem[Lee et~al.(2020)Lee, Kim, and Ro]{lee2019bman}
Sangmin Lee, Hak~Gu Kim, and Yong~Man Ro.
\newblock {BMAN:} bidirectional multi-scale aggregation networks for abnormal event detection.
\newblock \emph{{IEEE} Trans. Image Process.}, 29:\penalty0 2395--2408, 2020.

\bibitem[Li et~al.(2022)Li, Liu, and Jiao]{li2022self}
Shuo Li, Fang Liu, and Licheng Jiao.
\newblock Self-training multi-sequence learning with transformer for weakly supervised video anomaly detection.
\newblock In \emph{Proc. {AAAI} Conf. Artif. Intell.}, pages 1395--1403, 2022.

\bibitem[Liang et~al.(2023)Liang, Wu, Dai, Li, Zhao, Zhang, Zhang, Vajda, and Marculescu]{liang2023open}
Feng Liang, Bichen Wu, Xiaoliang Dai, Kunpeng Li, Yinan Zhao, Hang Zhang, Peizhao Zhang, Peter Vajda, and Diana Marculescu.
\newblock Open-vocabulary semantic segmentation with mask-adapted {CLIP}.
\newblock In \emph{Proc. {IEEE/CVF} Conf. Comput. Vis. Pattern Recognit.}, pages 7061--7070, 2023.

\bibitem[Lin et~al.(2024)Lin, Ding, Zhou, Peng, Zhao, Loy, and Zheng]{lin2024rethinking}
Kun{-}Yu Lin, Henghui Ding, Jiaming Zhou, Yi{-}Xing Peng, Zhilin Zhao, Chen~Change Loy, and Wei{-}Shi Zheng.
\newblock Rethinking clip-based video learners in cross-domain open-vocabulary action recognition.
\newblock \emph{arXiv:2403.01560}, 2024.

\bibitem[Liu et~al.(2018)Liu, Luo, Lian, and Gao]{liu2018future}
Wen Liu, Weixin Luo, Dongze Lian, and Shenghua Gao.
\newblock Future frame prediction for anomaly detection - {A} new baseline.
\newblock In \emph{Proc. {IEEE/CVF} Conf. Comput. Vis. Pattern Recognit.}, pages 6536--6545, 2018.

\bibitem[Liu et~al.(2024)Liu, Zhao, Han, Yi, Jiang, Wang, and Zhong]{liu2024pixel}
Wenxuan Liu, Shilei Zhao, Xiyu Han, Aoyu Yi, Kui Jiang, Zheng Wang, and Xian Zhong.
\newblock Pixel-refocused navigated tri-margin for semi-supervised action detection.
\newblock In \emph{Proc. ACM Int. Conf. Multimedia Workshop}, pages 23--31, 2024.

\bibitem[Liu et~al.(2021)Liu, Nie, Long, Zhang, and Li]{liu2021hybrid}
Zhian Liu, Yongwei Nie, Chengjiang Long, Qing Zhang, and Guiqing Li.
\newblock A hybrid video anomaly detection framework via memory-augmented flow reconstruction and flow-guided frame prediction.
\newblock In \emph{Proc. {IEEE/CVF} Int. Conf. Comput. Vis.}, pages 13568--13577, 2021.

\bibitem[Lu et~al.(2013)Lu, Shi, and Jia]{lu2013abnormal}
Cewu Lu, Jianping Shi, and Jiaya Jia.
\newblock Abnormal event detection at 150 {FPS} in {MATLAB}.
\newblock In \emph{Proc. {IEEE/CVF} Int. Conf. Comput. Vis.}, pages 2720--2727, 2013.

\bibitem[Lu et~al.(2019)Lu, Kumar, Nabavi, and Wang]{lu2019future}
Yiwei Lu, K.~Mahesh Kumar, Seyed~Shahabeddin Nabavi, and Yang Wang.
\newblock Future frame prediction using convolutional {VRNN} for anomaly detection.
\newblock In \emph{Proc. {IEEE} Int. Conf. Adv. Video Signal Based Surveill.}, pages 1--8, 2019.

\bibitem[Lv et~al.(2023)Lv, Yue, Sun, Luo, Cui, and Zhang]{lv2023unbiased}
Hui Lv, Zhongqi Yue, Qianru Sun, Bin Luo, Zhen Cui, and Hanwang Zhang.
\newblock Unbiased multiple instance learning for weakly supervised video anomaly detection.
\newblock In \emph{Proc. {IEEE/CVF} Conf. Comput. Vis. Pattern Recognit.}, pages 8022--8031, 2023.

\bibitem[OpenAI(2023)]{achiam2023gpt}
OpenAI.
\newblock {GPT-4} technical report.
\newblock \emph{arXiv:2303.08774}, 2023.

\bibitem[Radford et~al.(2021)Radford, Kim, Hallacy, Ramesh, Goh, Agarwal, Sastry, Askell, Mishkin, Clark, Krueger, and Sutskever]{radford2021learning}
Alec Radford, Jong~Wook Kim, Chris Hallacy, Aditya Ramesh, Gabriel Goh, Sandhini Agarwal, Girish Sastry, Amanda Askell, Pamela Mishkin, Jack Clark, Gretchen Krueger, and Ilya Sutskever.
\newblock Learning transferable visual models from natural language supervision.
\newblock In \emph{Proc. Int. Conf. Mach. Learn.}, pages 8748--8763, 2021.

\bibitem[Ristea et~al.(2024)Ristea, Croitoru, Ionescu, Popescu, Khan, and Shah]{ristea2024self}
Nicolae{-}Catalin Ristea, Florinel{-}Alin Croitoru, Radu~Tudor Ionescu, Marius Popescu, Fahad~Shahbaz Khan, and Mubarak Shah.
\newblock Self-distilled masked auto-encoders are efficient video anomaly detectors.
\newblock In \emph{Proc. {IEEE/CVF} Conf. Comput. Vis. Pattern Recognit.}, pages 15984--15995, 2024.

\bibitem[Sabokrou et~al.(2018)Sabokrou, Khalooei, Fathy, and Adeli]{sabokrou2018adversarially}
Mohammad Sabokrou, Mohammad Khalooei, Mahmood Fathy, and Ehsan Adeli.
\newblock Adversarially learned one-class classifier for novelty detection.
\newblock In \emph{Proc. {IEEE/CVF} Conf. Comput. Vis. Pattern Recognit.}, pages 3379--3388, 2018.

\bibitem[Sch{\"{o}}lkopf et~al.(1999)Sch{\"{o}}lkopf, Williamson, Smola, Shawe{-}Taylor, and Platt]{scholkopf1999support}
Bernhard Sch{\"{o}}lkopf, Robert~C. Williamson, Alexander~J. Smola, John Shawe{-}Taylor, and John~C. Platt.
\newblock Support vector method for novelty detection.
\newblock In \emph{Adv. Neural Inf. Process. Syst.}, pages 582--588, 1999.

\bibitem[Sultani et~al.(2018)Sultani, Chen, and Shah]{sultani2018real}
Waqas Sultani, Chen Chen, and Mubarak Shah.
\newblock Real-world anomaly detection in surveillance videos.
\newblock In \emph{Proc. {IEEE/CVF} Conf. Comput. Vis. Pattern Recognit.}, pages 6479--6488, 2018.

\bibitem[Tian et~al.(2021)Tian, Pang, Chen, Singh, Verjans, and Carneiro]{tian2021weakly}
Yu Tian, Guansong Pang, Yuanhong Chen, Rajvinder Singh, Johan~W. Verjans, and Gustavo Carneiro.
\newblock Weakly-supervised video anomaly detection with robust temporal feature magnitude learning.
\newblock In \emph{Proc. {IEEE/CVF} Int. Conf. Comput. Vis.}, pages 4955--4966, 2021.

\bibitem[Wang and Cherian(2019)]{wang2019gods}
Jue Wang and Anoop Cherian.
\newblock {GODS:} generalized one-class discriminative subspaces for anomaly detection.
\newblock In \emph{Proc. {IEEE/CVF} Int. Conf. Comput. Vis.}, pages 8200--8210, 2019.

\bibitem[Wang et~al.(2021)Wang, Xing, and Liu]{wang2021actionclip}
Mengmeng Wang, Jiazheng Xing, and Yong Liu.
\newblock Actionclip: {A} new paradigm for video action recognition.
\newblock \emph{arXiv:2109.08472}, 2021.

\bibitem[Wasim et~al.(2024)Wasim, Naseer, Khan, Yang, and Khan]{wasim2024videogrounding}
Syed~Talal Wasim, Muzammal Naseer, Salman~H. Khan, Ming{-}Hsuan Yang, and Fahad~Shahbaz Khan.
\newblock Videogrounding-dino: Towards open-vocabulary spatio- temporal video grounding.
\newblock In \emph{Proc. {IEEE/CVF} Conf. Comput. Vis. Pattern Recognit.}, pages 18909--18918, 2024.

\bibitem[Weng et~al.(2023)Weng, Yang, Li, Wu, and Jiang]{weng2023open}
Zejia Weng, Xitong Yang, Ang Li, Zuxuan Wu, and Yu{-}Gang Jiang.
\newblock Open-vclip: Transforming {CLIP} to an open-vocabulary video model via interpolated weight optimization.
\newblock In \emph{Proc. Int. Conf. Mach. Learn.}, pages 36978--36989, 2023.

\bibitem[Wu and Liu(2021)]{wu2021learning}
Peng Wu and Jing Liu.
\newblock Learning causal temporal relation and feature discrimination for anomaly detection.
\newblock \emph{{IEEE} Trans. Image Process.}, 30:\penalty0 3513--3527, 2021.

\bibitem[Wu et~al.(2020{\natexlab{a}})Wu, Liu, and Shen]{wu2019deep}
Peng Wu, Jing Liu, and Fang Shen.
\newblock A deep one-class neural network for anomalous event detection in complex scenes.
\newblock \emph{{IEEE} Trans. Neural Networks Learn. Syst.}, 31\penalty0 (7):\penalty0 2609--2622, 2020{\natexlab{a}}.

\bibitem[Wu et~al.(2020{\natexlab{b}})Wu, Liu, Shi, Sun, Shao, Wu, and Yang]{wu2020not}
Peng Wu, Jing Liu, Yujia Shi, Yujia Sun, Fangtao Shao, Zhaoyang Wu, and Zhiwei Yang.
\newblock Not only look, but also listen: Learning multimodal violence detection under weak supervision.
\newblock In \emph{Proc. Eur. Conf. Comput. Vis.}, pages 322--339, 2020{\natexlab{b}}.

\bibitem[Wu et~al.(2024{\natexlab{a}})Wu, Zhou, Pang, Sun, Liu, Wang, and Zhang]{wu2024open}
Peng Wu, Xuerong Zhou, Guansong Pang, Yujia Sun, Jing Liu, Peng Wang, and Yanning Zhang.
\newblock Open-vocabulary video anomaly detection.
\newblock In \emph{Proc. {IEEE/CVF} Conf. Comput. Vis. Pattern Recognit.}, pages 18297--18307, 2024{\natexlab{a}}.

\bibitem[Wu et~al.(2024{\natexlab{b}})Wu, Zhou, Pang, Zhou, Yan, Wang, and Zhang]{wu2024vadclip}
Peng Wu, Xuerong Zhou, Guansong Pang, Lingru Zhou, Qingsen Yan, Peng Wang, and Yanning Zhang.
\newblock Vadclip: Adapting vision-language models for weakly supervised video anomaly detection.
\newblock In \emph{Proc. {AAAI} Conf. Artif. Intell.}, pages 6074--6082, 2024{\natexlab{b}}.

\bibitem[Wu et~al.(2024{\natexlab{c}})Wu, Ge, Qin, Wu, and Wang]{wu2024open2}
Tao Wu, Shuqiu Ge, Jie Qin, Gangshan Wu, and Limin Wang.
\newblock Open-vocabulary spatio-temporal action detection.
\newblock \emph{arXiv:2405.10832}, 2024{\natexlab{c}}.

\bibitem[Wu et~al.(2024{\natexlab{d}})Wu, Weng, Peng, Yang, Li, Davis, and Jiang]{wu2024building}
Zuxuan Wu, Zejia Weng, Wujian Peng, Xitong Yang, Ang Li, Larry~S. Davis, and Yu{-}Gang Jiang.
\newblock Building an open-vocabulary video {CLIP} model with better architectures, optimization and data.
\newblock \emph{{IEEE} Trans. Pattern Anal. Mach. Intell.}, 46\penalty0 (7):\penalty0 4747--4762, 2024{\natexlab{d}}.

\bibitem[Xie et~al.(2023)Xie, Yang, Zhu, and Wang]{xie2023striking}
Haiyang Xie, Zhengwei Yang, Huilin Zhu, and Zheng Wang.
\newblock Striking a balance: Unsupervised cross-domain crowd counting via knowledge diffusion.
\newblock In \emph{Proceedings of the 31st ACM international conference on multimedia}, pages 6520--6529, 2023.

\bibitem[Xu et~al.(2023)Xu, Zhang, Wei, Hu, and Bai]{xu2023side}
Mengde Xu, Zheng Zhang, Fangyun Wei, Han Hu, and Xiang Bai.
\newblock {SAN:} side adapter network for open-vocabulary semantic segmentation.
\newblock \emph{{IEEE} Trans. Pattern Anal. Mach. Intell.}, 45\penalty0 (12):\penalty0 15546--15561, 2023.

\bibitem[Yang et~al.(2023)Yang, Liu, Wu, Wu, and Liu]{yang2023video}
Zhiwei Yang, Jing Liu, Zhaoyang Wu, Peng Wu, and Xiaotao Liu.
\newblock Video event restoration based on keyframes for video anomaly detection.
\newblock In \emph{Proc. {IEEE/CVF} Conf. Comput. Vis. Pattern Recognit.}, pages 14592--14601, 2023.

\bibitem[Zaheer et~al.(2020{\natexlab{a}})Zaheer, Lee, Astrid, and Lee]{zaheer2020old}
Muhammad~Zaigham Zaheer, Jin{-}Ha Lee, Marcella Astrid, and Seung{-}Ik Lee.
\newblock Old is gold: Redefining the adversarially learned one-class classifier training paradigm.
\newblock In \emph{Proc. {IEEE/CVF} Conf. Comput. Vis. Pattern Recognit.}, pages 14171--14181, 2020{\natexlab{a}}.

\bibitem[Zaheer et~al.(2020{\natexlab{b}})Zaheer, Mahmood, Astrid, and Lee]{zaheer2020claws}
Muhammad~Zaigham Zaheer, Arif Mahmood, Marcella Astrid, and Seung{-}Ik Lee.
\newblock {CLAWS:} clustering assisted weakly supervised learning with normalcy suppression for anomalous event detection.
\newblock In \emph{Proc. Eur. Conf. Comput. Vis.}, pages 358--376, 2020{\natexlab{b}}.

\bibitem[Zareian et~al.(2021)Zareian, Rosa, Hu, and Chang]{zareian2021open}
Alireza Zareian, Kevin~Dela Rosa, Derek~Hao Hu, and Shih{-}Fu Chang.
\newblock Open-vocabulary object detection using captions.
\newblock In \emph{Proc. {IEEE/CVF} Conf. Comput. Vis. Pattern Recognit.}, pages 14393--14402, 2021.

\bibitem[Zhong et~al.(2022)Zhong, Chen, Jiang, and Ren]{zhong2022cascade}
Yuanhong Zhong, Xia Chen, Jinyang Jiang, and Fan Ren.
\newblock A cascade reconstruction model with generalization ability evaluation for anomaly detection in videos.
\newblock \emph{Pattern Recognit.}, 122:\penalty0 108336, 2022.

\bibitem[Zhou et~al.(2022)Zhou, Yang, Loy, and Liu]{zhou2022learning}
Kaiyang Zhou, Jingkang Yang, Chen~Change Loy, and Ziwei Liu.
\newblock Learning to prompt for vision-language models.
\newblock \emph{Int. J. Comput. Vis.}, 130\penalty0 (9):\penalty0 2337--2348, 2022.

\bibitem[Zhu et~al.(2022)Zhu, Bao, and Yu]{zhu2022towards}
Yuansheng Zhu, Wentao Bao, and Qi Yu.
\newblock Towards open set video anomaly detection.
\newblock In \emph{Proc. Eur. Conf. Comput. Vis.}, pages 395--412, 2022.

\end{thebibliography}
}


\clearpage
\setcounter{page}{1}
\setcounter{section}{0}
\maketitlesupplementary

% \noindent \paragraph{1. Overview}
\section{Overview}

% \noindent 
This supplementary material is organized as follows:
\begin{itemize}
\item Data Division
\item Impact of Data Addition
\item Impact of Using Other Lightweight Temporal Encoder
\item t-SNE Visualization of Video Features
\item Visualization of Per-Class Results
\item Prompt Design
\item Limitation
\end{itemize}

\section{Data Division}

% \noindent 
In the OVVAD task, we categorize anomaly labels into base and novel categories, utilizing only base category samples during training. Following standard open vocabulary learning practices, frequent labels are designated as base categories, while rare labels are classified as novel categories. Specifically, for \textsc{UCF-Crime} dataset, the base categories include Abuse, Assault, Burglary, Road Accident, Robbery, and Stealing, with all other labels treated as novel. On \textsc{XD-Violence} dataset, the base categories are Fighting, Shooting, and Car Accident.

\section{Impact of Data Addition}

\cref{table:afterAddingData_performance} indicates that the performance drop after adding new labels is acceptable. These new labels consist of both same-group and cross-group categories. Same-group labels can lead to closer encodings, potentially interfering with original predictions and slightly affecting performance. For \textsc{XD-Violence} and \textsc{UCF-Crime}, the added overlapping categories include drug trafficking, harassment, stalking, loitering, and public intoxication. Additionally, \textsc{XD-Violence} incorporates shoplifting, arson, robbery, and arrest, while \textsc{UCF-Crime} includes riot.

We provide a detailed analysis of the results presented in \cref{table:afterAddingData_performance}. After adding new labels and data, the model correctly categorizes 20 of 21 shoplifting samples, 3 of 5 arrest samples, 7 of 10 arson samples, 3 of 3 assault samples, and 91 of 99 riot samples. Among these, only shoplifting is not grouped with the base labels. For anomalous data within the same group of labels, the similarity in visual encodings suggests a higher risk of overfitting, potentially leading to miscategorization as base cases within the same group. Nonetheless, our method achieves excellent results, highlighting the effectiveness of the guided text encodings in reducing categorization confusion.

\section{Impact of Using Other Lightweight Temporal Encoder}

% \begin{table}[htbp]
%	\centering
%	\resizebox{0.45\textwidth}{!}{ % 调整表格宽度
%	\begin{tabular}{lccc}
%	\toprule
%	\textbf{Dataset} & \textbf{ACC (\%)} & \textbf{ACC$_b$ (\%)} & \textbf{ACC$_n$ (\%)} \\
%	\midrule
%	 \textsc{XD-Violence} & 83.89 & 95.04 & 68.59 \\
%	\textsc{UCF-Crime} & 46.43 & 52.94 & 42.70 \\
%	\bottomrule
%	\end{tabular}
%	\vspace{-5cm}
%	 \caption{\textbf{Top-1 Accuracy of Using Transformer as the Lightweight Temporal Encoder.}}
% \end{table}

% \noindent 
We implement an LSTM as a lightweight temporal encoder to mitigate overfitting to base classes and reduce label confusion in novel cases. This design choice is motivated by our observation that highly parameterized encoders are prone to overfitting, likely because the additional parameters tend to capture features that closely resemble those in the training data when handling novel data. As a result, when these visual features align with the label textual features, they are found to be closer to the labels of the base data, thereby leading to misclassification.

In the supplementary material, we also evaluate a transformer to validate our method, with results presented in \cref{table:transformer_performance}. However, due to having more parameters, the transformer underperformed compared to the LSTM on both datasets. On \textsc{UCF-Crime}, the transformer required a lower learning rate of $5 \times 10^{-8}$ to prevent overfitting. While this adjustment improves performance for novel classes, it significantly reduces performance for base classes, resulting in poorer overall results. Regardless of the temporal encoder employed, our method consistently outperforms existing SOTA models, indirectly confirming the effectiveness of our group-guided text encoding mechanism.

\begin{table}[h]
	% \vspace{-8pt}
	\centering
	\footnotesize
	\setlength{\tabcolsep}{10pt}
	\begin{tabular}{lccc}
	\toprule[1.1pt]
	{Dataset} & {ACC} & {ACC$_b$} & {ACC$_n$} \\
	\midrule
	\textsc{XD-Violence} & 83.89 & 95.04 & 68.59 \\
	\textsc{UCF-Crime} & 46.43 & 52.94 & 42.70 \\
	\bottomrule[1.1pt]
	\end{tabular}
	\vspace{-5pt}
	\caption{\textbf{Top-1 Accuracy (\%) with a Transformer-Based Lightweight Temporal Encoder.}}
	\label{table:transformer_performance}
	\vspace{-5pt}
\end{table}

\section{t-SNE Visualization of Video Features}

\begin{figure*}
	\centering
	\begin{tabular}{cccc}
	
	\subfloat[Original Visual Feature \label{fig:xd-origin-feature}]{
	\includegraphics[width = 0.2\linewidth]{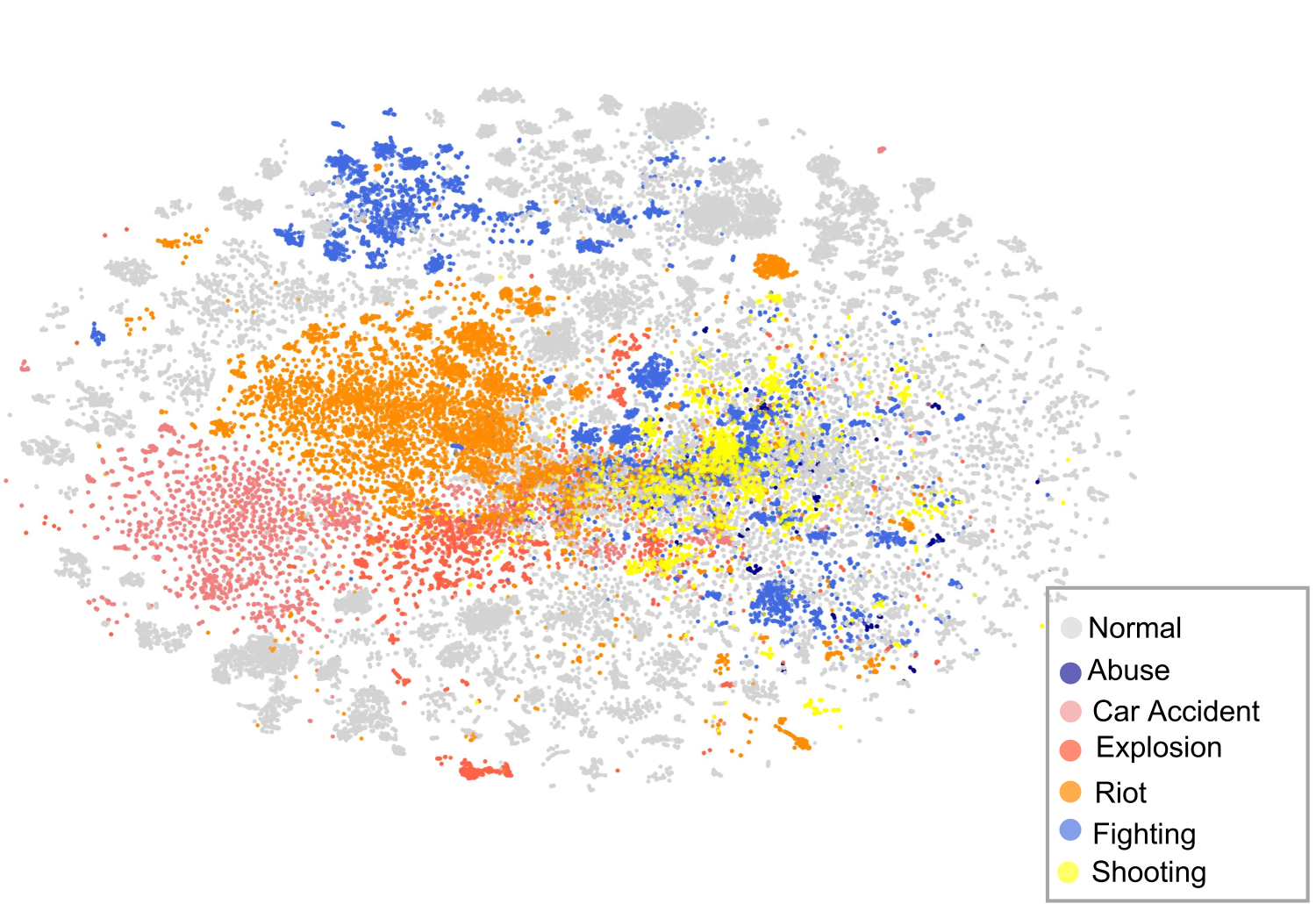}
	} & 
	\subfloat[Static Stream Visual Feature \label{fig:xd-static-feature}]{
	\includegraphics[width = 0.2\linewidth]{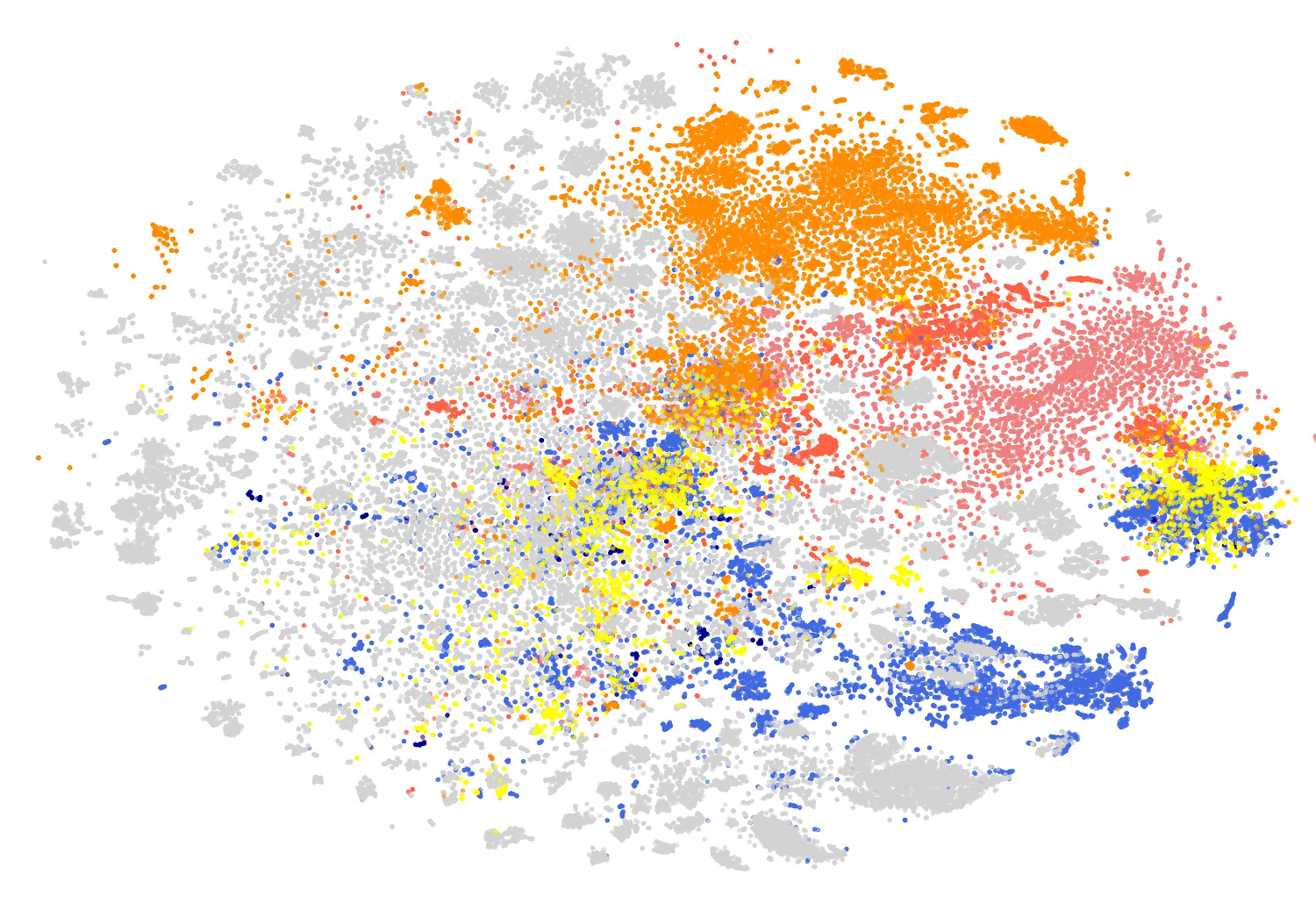}
	} & 
	\subfloat[Dynamic Stream Visual Feature \label{fig:xd-dynamic-feature}]{
	\includegraphics[width = 0.2\linewidth]{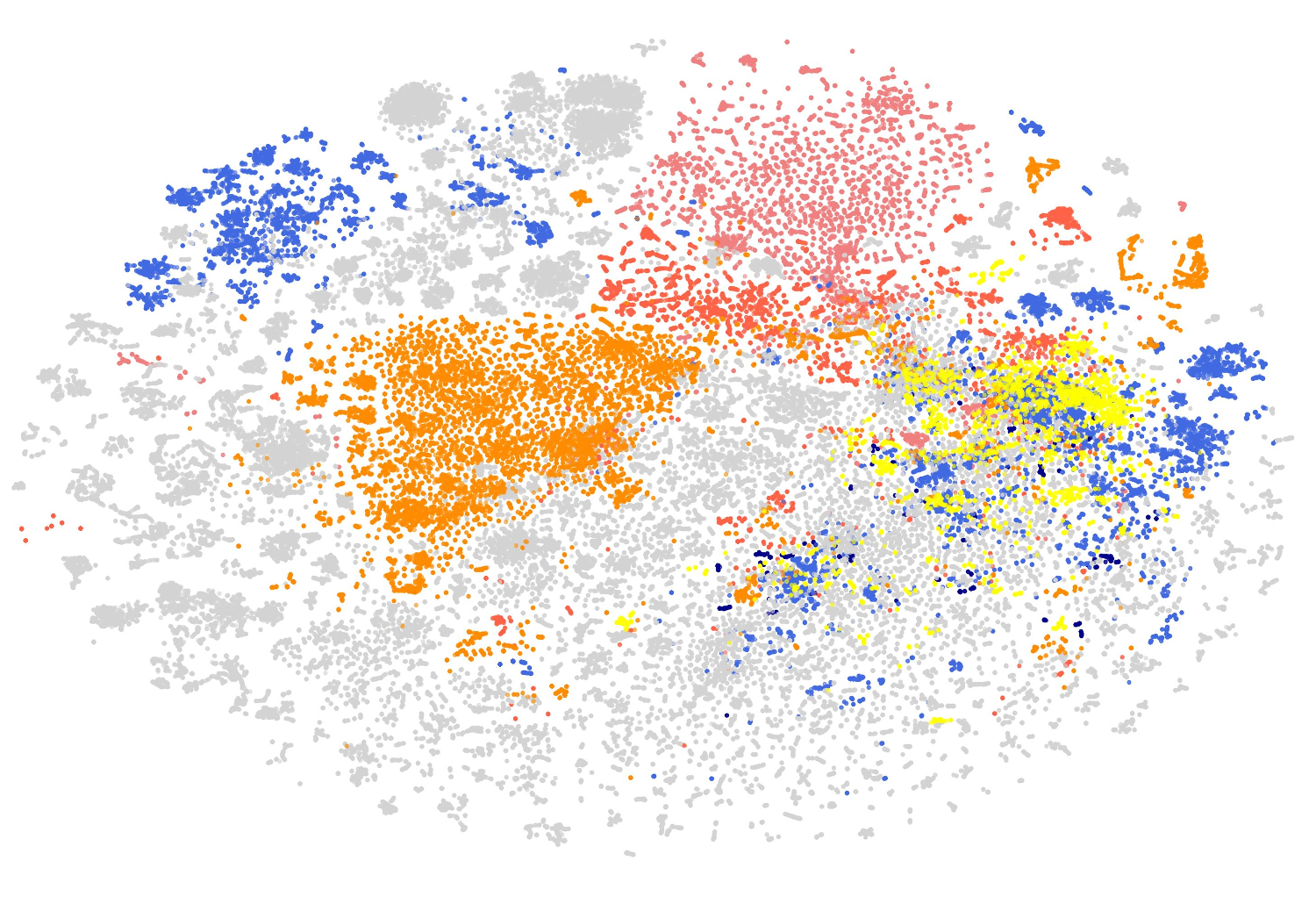}
	} & 
	\subfloat[Visual Feature for Categorization \label{fig:xd-categorization-feature}]{
	\includegraphics[width = 0.2\linewidth]{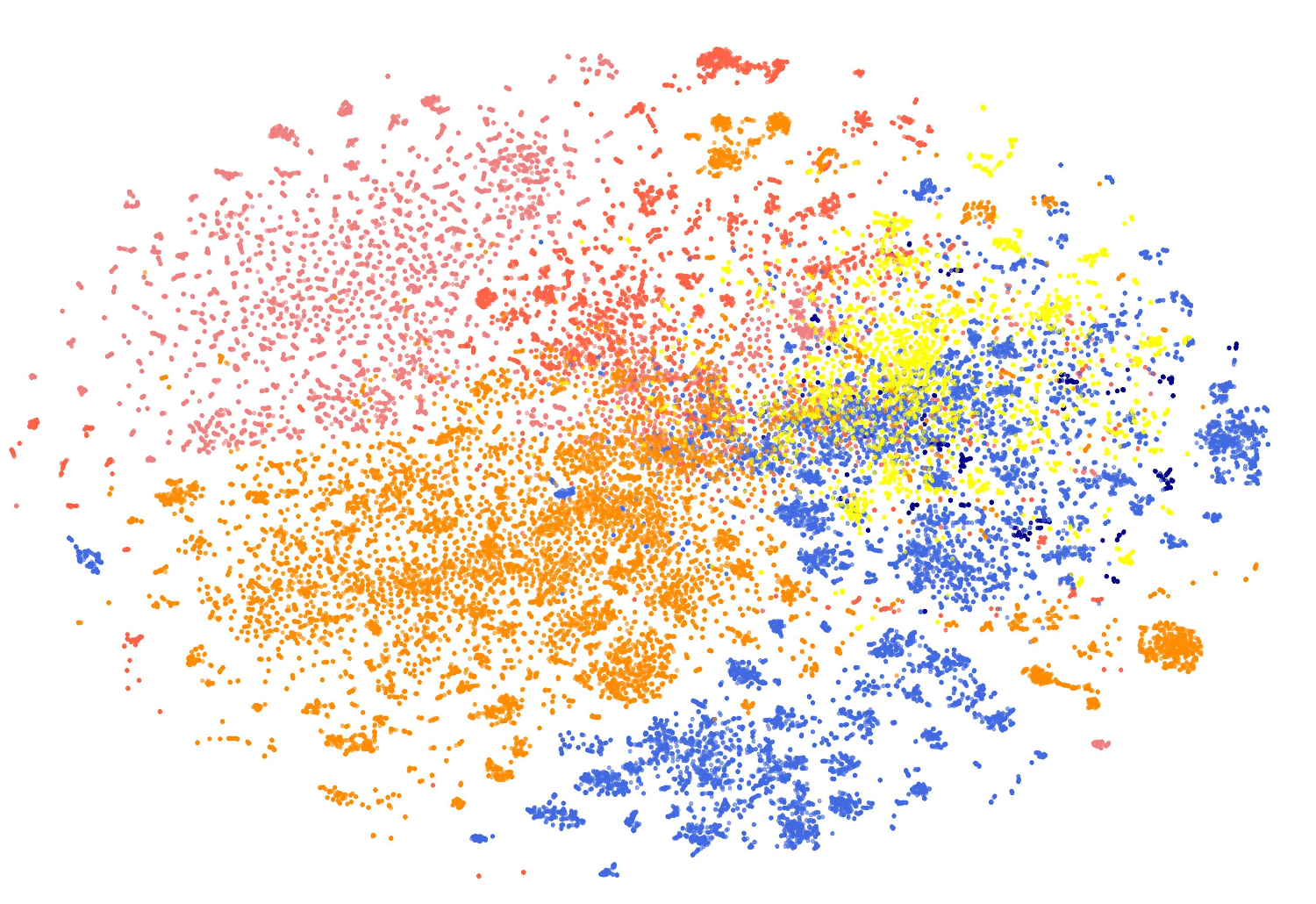}
	} \\
	
	\subfloat[Original Visual Feature \label{fig:ucf-origin-feature}]{
	\includegraphics[width = 0.2\linewidth]{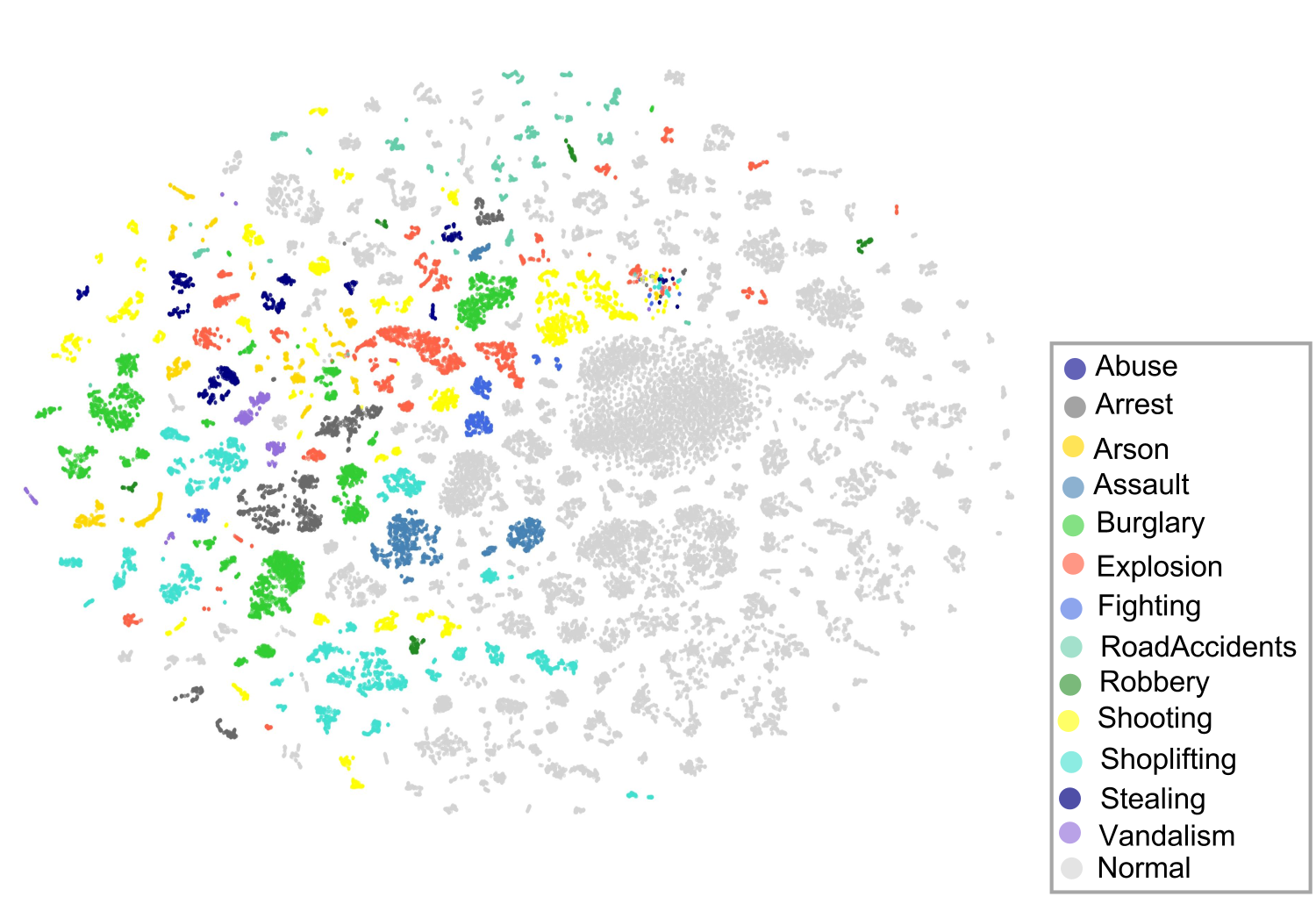}
	} & 
	\subfloat[Static Stream Visual Feature \label{fig:ucf-static-feature}]{
	\includegraphics[width = 0.2\linewidth]{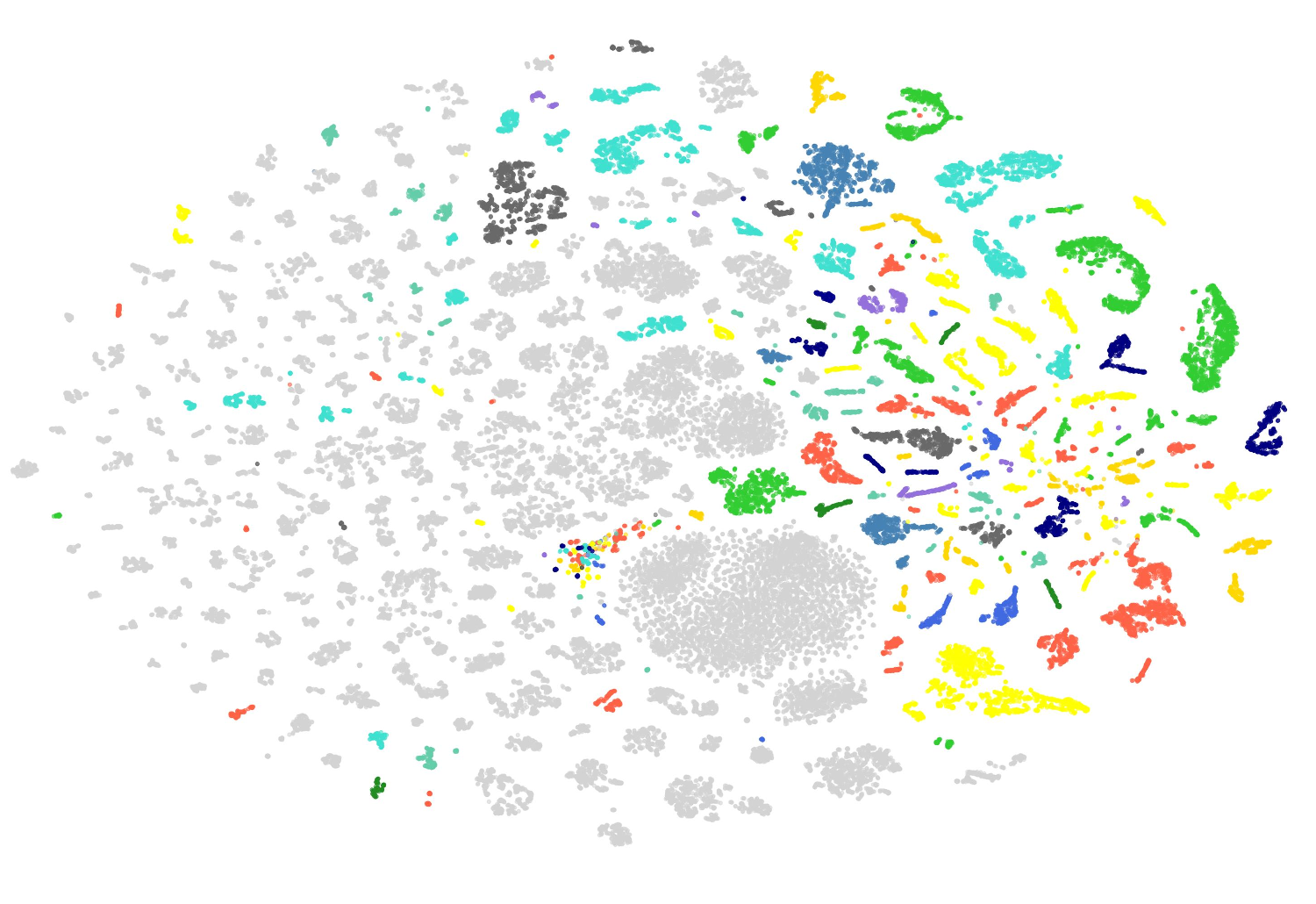}
	} & 
	\subfloat[Dynamic Stream Visual Feature \label{fig:ucf-dynamic-feature}]{
	\includegraphics[width = 0.2\linewidth]{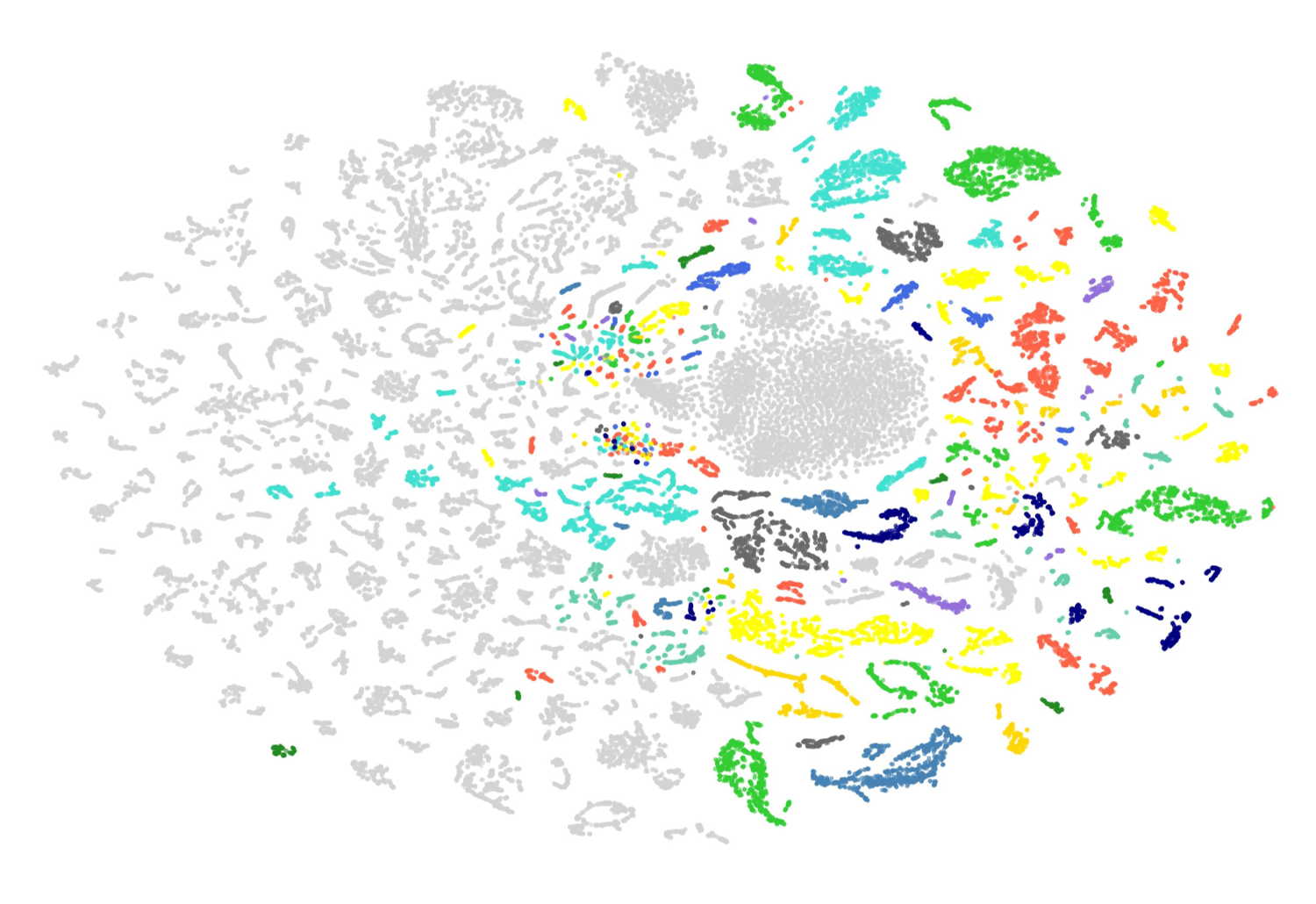}
	} & 
	\subfloat[Visual Feature for Categorization \label{fig:ucf-categorization-feature}]{
	\includegraphics[width = 0.2\linewidth]{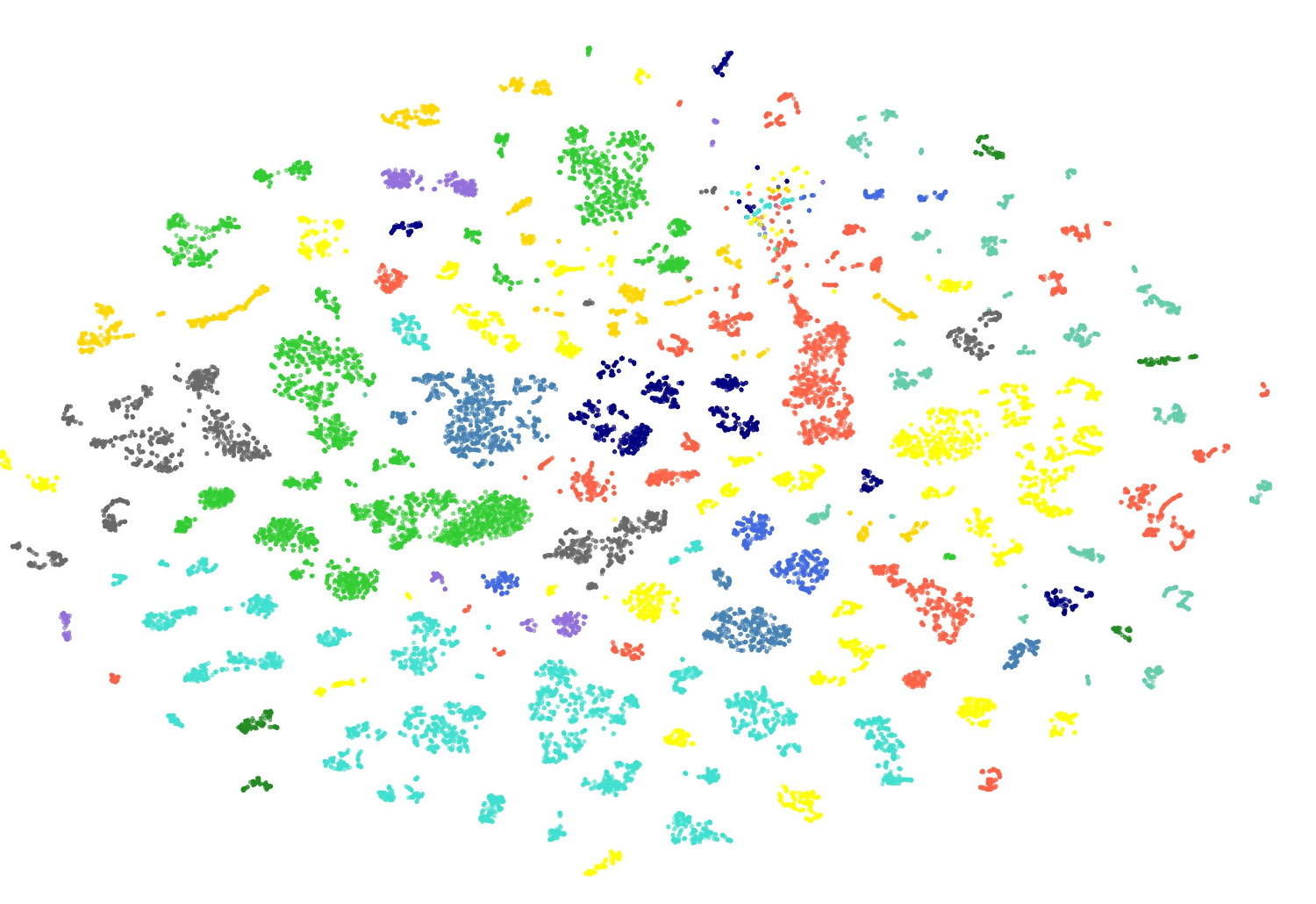}
	} \\
	\end{tabular}
	\vspace{-5pt}
	\caption{\textbf{Scatter Plots of Features Reduced to 2D Using t-SNE.} (a--d) correspond to \textsc{XD-Violence}, while (e--h) represent \textsc{UCF-Crime}. ``Original Visual Feature'' refers to scatter plots generated from CLIP frame encodings. ``Static Stream Visual Feature'' and ``Dynamic Stream Visual Feature'' denote features refined through the static and dynamic streams for detection, respectively. ``Visual Feature for Categorization'' represents the fused visual features used for classification.}
	\label{fig:t-SNE_visualization}
	\vspace{-5pt}
\end{figure*}

% \noindent 
As shown in \cref{fig:t-SNE_visualization}, the original CLIP encodings exhibit irregular patterns, making them unsuitable for anomaly detection and categorization. In contrast, \cref{fig:t-SNE_visualization}(b), (c), (f), and (g) demonstrate that the text-augmented static and dynamic streams provide sufficient information, resulting in clearer boundaries between normal and anomalous data. This facilitates the detector in better distinguishing between them.
After temporal modeling and feature fusion, as shown in \cref{fig:t-SNE_visualization}(d) and (h), more distinct clusters emerge, with data points sharing the same label becoming closer in high-dimensional space, thereby enhancing categorization.
Notably, the visualization reflects the frame-level features used for detection and categorization, and the presence of normal frames within anomalous videos contributes to minor overlaps.

\begin{figure*}
	\centering
	\begin{tabular}{cc}
	\includegraphics[width = 0.45\linewidth]{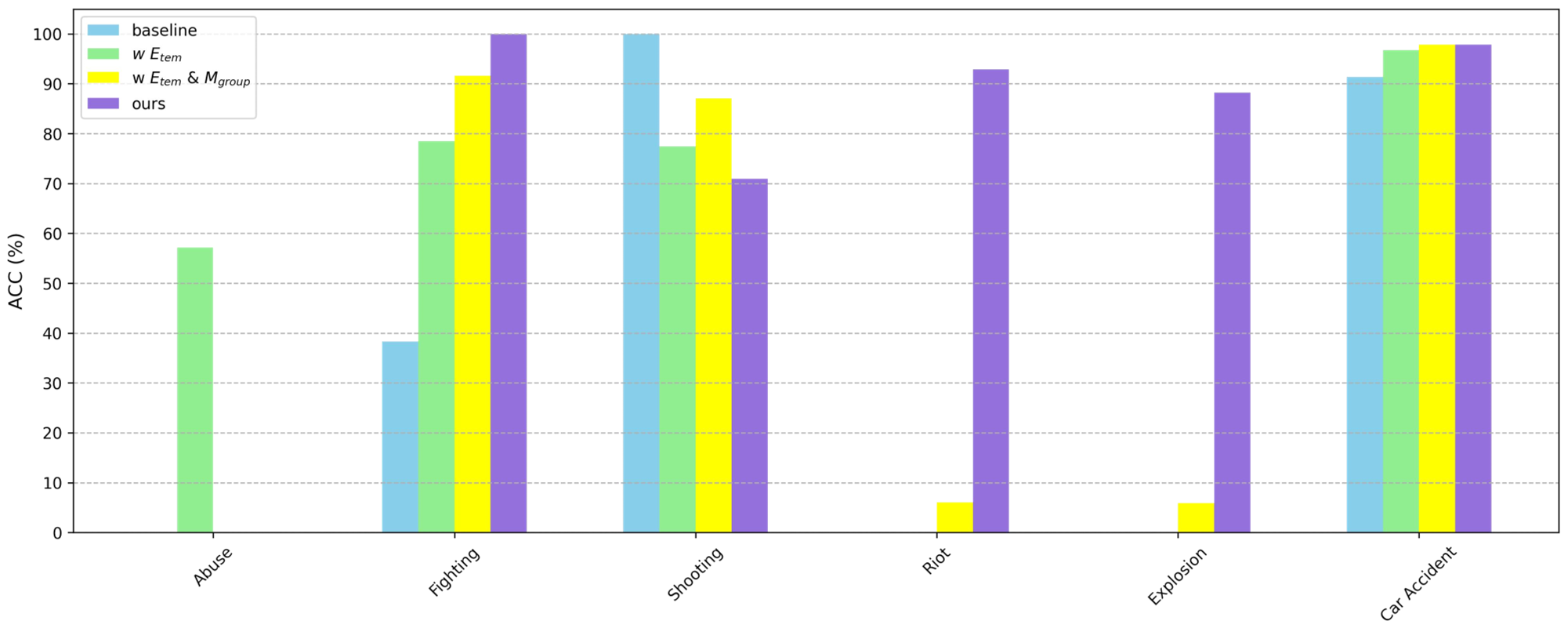} & 
	\includegraphics[width = 0.45\linewidth]{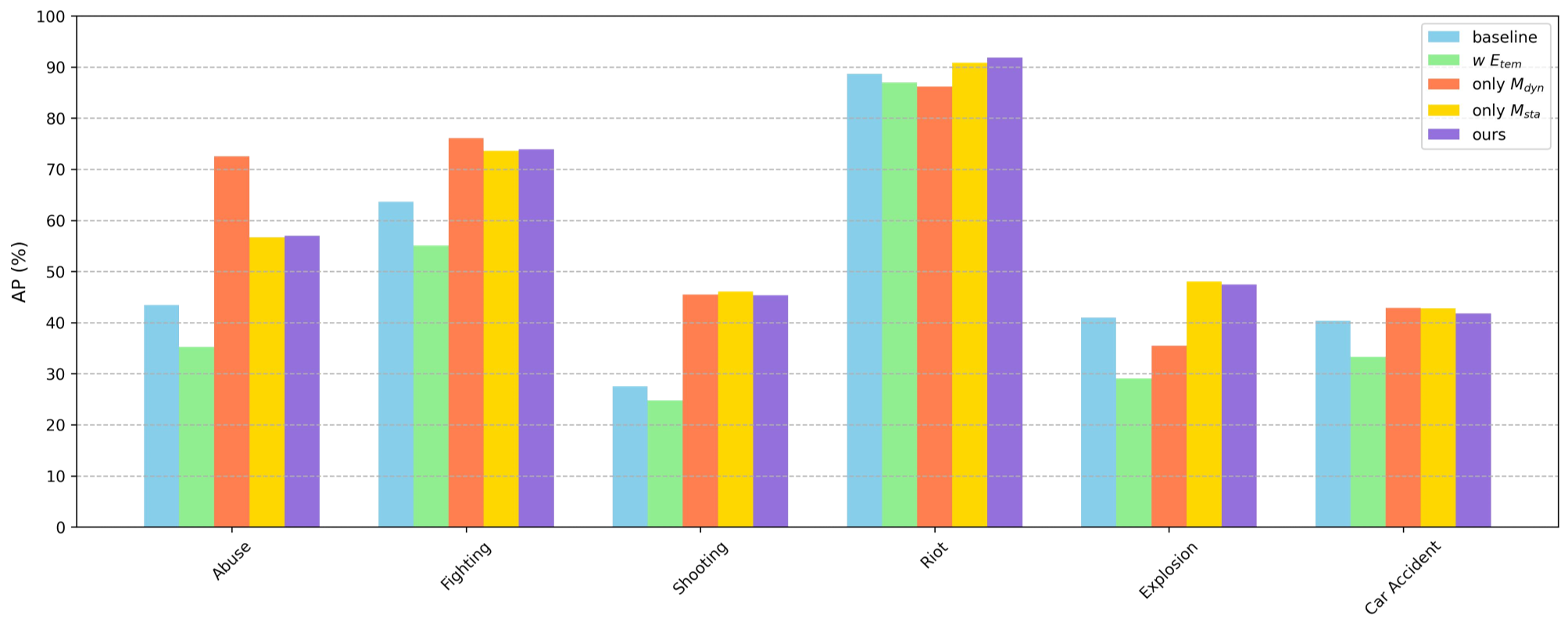} \\
	\footnotesize{(a) Per-class Detection Results on \textsc{XD-Violence}} & 
	\footnotesize{(b) Per-class Categorization Results on \textsc{XD-Violence}} \\
	
	\includegraphics[width = 0.45\linewidth]{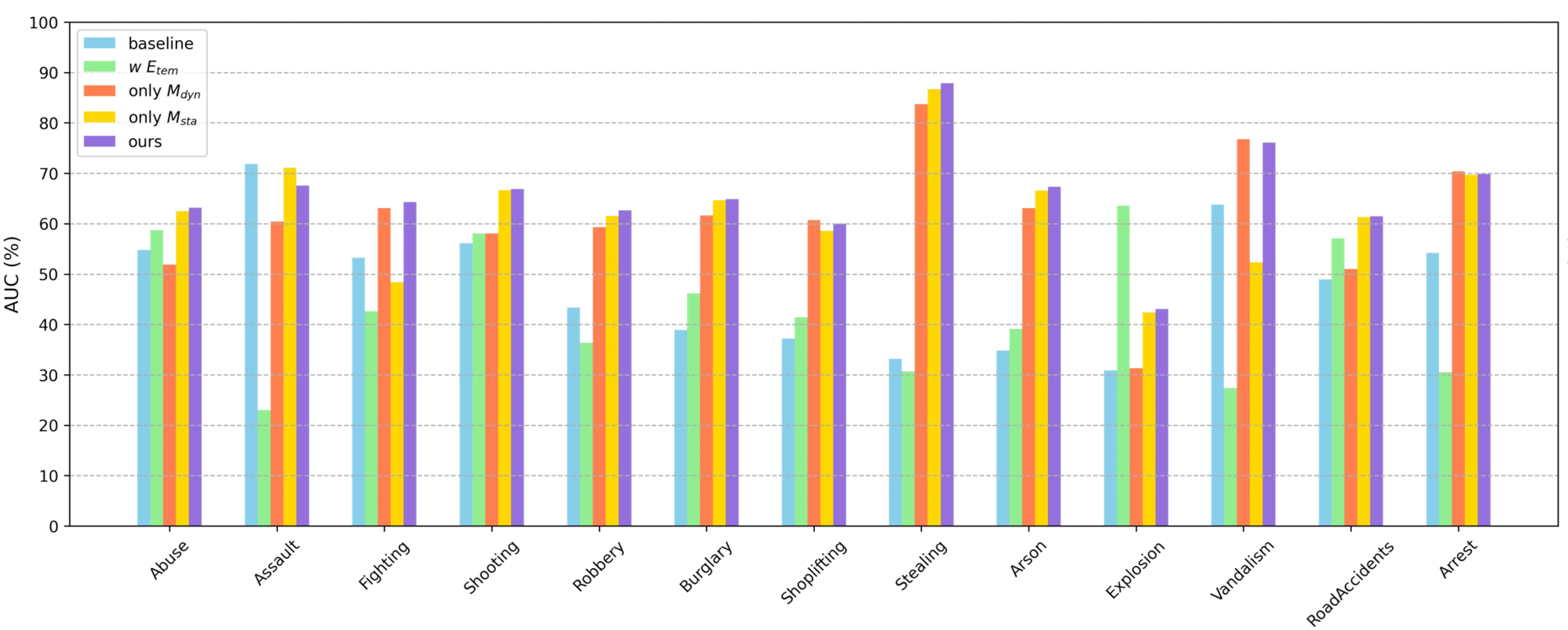} & 
	\includegraphics[width = 0.45\linewidth]{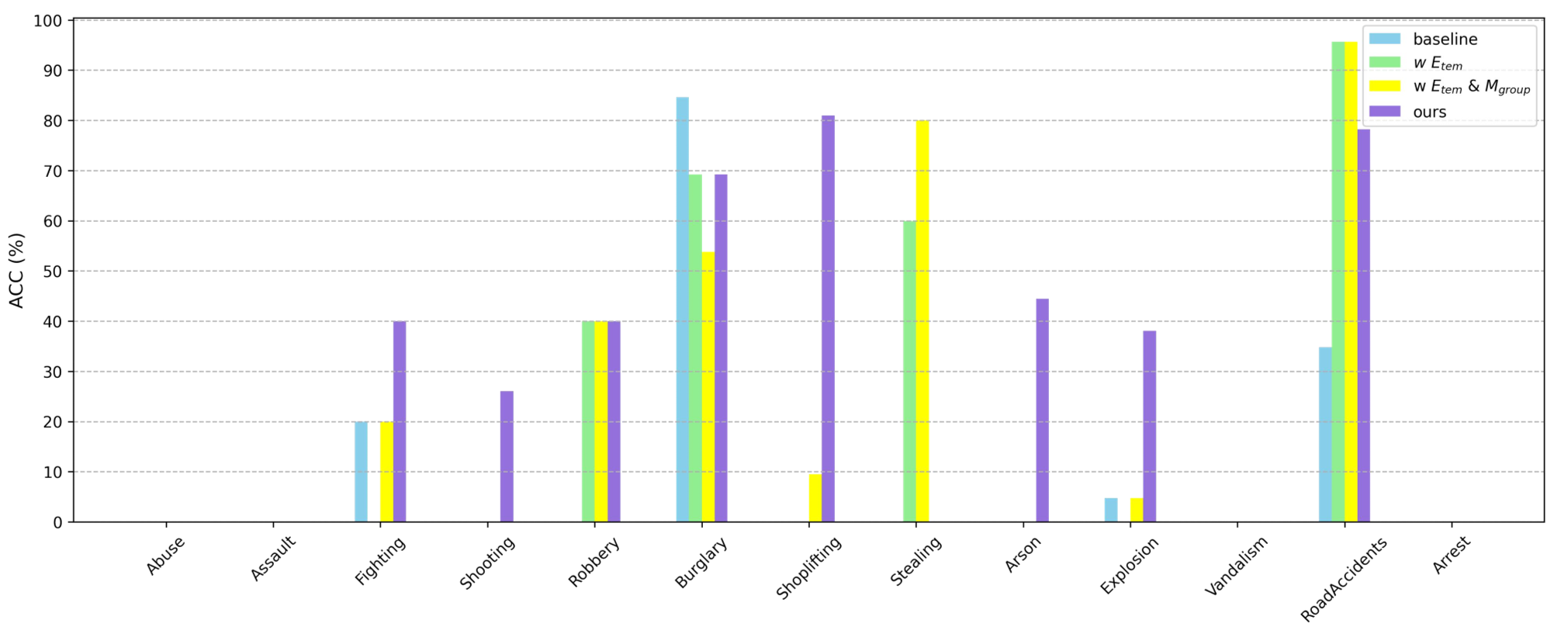} \\
	\footnotesize{(c) Per-class Detection Results on \textsc{UCF-Crime}} & 
	\footnotesize{(d) Per-class Categorization Results on \textsc{UCF-Crime}} \\
	\end{tabular}
	\vspace{-5pt}
	\caption{\textbf{Per-Class Contribution of Each Design Element to Open Vocabulary Anomaly Detection on \textsc{XD-Violence} and \textsc{UCF-Crime}.} The baseline uses original CLIP encodings, while ``w $E_\mathrm{tem}$'' incorporates temporal encoding. ``Only $M_\mathrm{dyn}$'' and ``Only $M_\mathrm{sta}$'' denote the dynamic and static streams, respectively, with text augmentation and loss weight $w_i$. ``w $E_\mathrm{tem}$ \& $M_\mathrm{group}$'' further integrates the guided encoding mechanism.}
	\label{fig:per-class-result}
	\vspace{-5pt}
\end{figure*}

\section{Visualization of Per-Class Results}

% \noindent 
\cref{fig:per-class-result} illustrates the impact of key design elements on performance across different categories. In both categorization and detection branches, our method achieves the best or near-best results across most categories, demonstrating that our approach not only ensures strong performance on base anomalies but also effectively addresses novel anomalies in OVVAD.

% \noindent Fig. \ref{fig:per-class-result} shows the impact of designs on performance across different categories. 
% Fig. \ref{fig:UCF-perclass-detection} and \ref{fig:XD-perclass-detection} present detection performance.
% Our method combines two streams and achieves the best or near-best performance across datasets, validating its effectiveness in detection.
% Fig. \ref{fig:UCF-perclass-Categorizaion} and \ref{fig:XD-perclass-Categorizaion} present categorization performance.
% Our method achieving best or near-best results in most cases and demonstrating strong overall performance.

% 描述prompt的设计，怎么才能更好的完成我们的目标
\section{Prompt Design}

% \noindent 
We provide the prompt designs described in the paper to offer deeper insights into our method and explain why the group-guided text encoding mechanism and $ConceptLib$ designs lead to significant performance improvements. After generating group data using $\mathrm{prompt}_\mathrm{group}$, we manually refine the results to eliminate noise. For $\mathrm{prompt}_\mathrm{desc}$, we input a group of labels with length constraints to align with the CLIP text encoder requirements. The value of $L$ in $\mathrm{prompt}_\mathrm{conc}$ is determined by the number of labels: \textsc{XD-Violence}, with six anomaly labels, uses $L = 200$, while \textsc{UCF-Crime}, with thirteen labels, uses $L = 500$.

% \noindent 
\vspace{-10pt}
\paragraph{$\mathrm{prompt}_\mathrm{group}$.} Group the following anomaly labels: \{labels\}. Organize them based on similarities in visual characteristics, where behaviors with comparable actions, activities, or scene contexts during the anomaly are placed in the same group.

% \noindent 
\vspace{-10pt}
\paragraph{$\mathrm{prompt}_\mathrm{desc}$.} Describe the anomaly of \{labels\}. Begin each description with one or two sentences that emphasize the common traits shared among all behaviors, followed by one or two sentences detailing the unique characteristics specific to each behavior. Ensure the descriptions clearly capture significant details of the anomaly, such as actions, movements, and scene context. Maintain a consistent sentence structure for each description, with word counts between 50 and 70 words.

% \noindent
\vspace{-10pt}
\paragraph{$\mathrm{prompt}_\mathrm{conc}$.} Given the anomaly labels: \{labels\}, generate \{$L$\} noun phrases that accurately capture the key scene characteristics associated with each label. These phrases should be well-suited for CLIP model encoding and designed to complement visual features, enhancing the effectiveness of model in anomaly detection.

\section{Limitation}
In this paper, we distinguish dynamic and static elements at the semantic level, considering dynamic elements as label descriptions that appear in the form of sentences, and static elements as nouns that describe key features of the anomaly. There may be slight redundancy between these two elements, but it does not affect our ability to achieve outstanding performance. However, exploring feature-level disentanglement for them may further enhance performance.

\end{document}